%% file: acl_latex.tex
\documentclass[11pt]{article}

\usepackage[final]{acl}
\usepackage{etoolbox}

\newtoggle{anonymise}
\togglefalse{anonymise}

\makeatletter
\@ifpackagewith{acl}{review}{
  \toggletrue{anonymise}
}{
  \togglefalse{anonymise}
}
\makeatother

\usepackage{rotating}

\input{def}

\usepackage{times}
\usepackage{latexsym}

\usepackage[T1]{fontenc}

\usepackage[utf8]{inputenc}

\usepackage{microtype}

\usepackage{inconsolata}

\usepackage{graphicx}

\title{\ourdataset: A Multilingual Benchmark for Long-Form LLM-Generated Text Attribution under Distribution Shifts}

\author{
  Matteo Greco$^{1,2*}$,  Anudeex Shetty$^{1*}$, Andrea Tagarelli$^{2}$, and Jey Han Lau$^{1}$ \\
  $^1${School of Computing and Information Systems, The University of Melbourne, Australia} \\
  $^2${DIMES Dept., University of Calabria, Italy} \\
  {\tt matteogre01@gmail.com}, \\
  {\tt \{anudeex, laujh\}@unimelb.edu.au}, {\tt tagarelli@dimes.unical.it}
}

\begin{document}
\maketitle
\def\thefootnote{*}\footnotetext{Equal contributions.}\def\thefootnote{\arabic{footnote}} 
\begin{abstract}
While existing work on LLM authorship attribution (AA) has made progress, available benchmarks remain limited, often focusing on English, controlled settings, or relatively outdated models, {\color{black}with the few multilingual studies considering only relatively short texts.}  We introduce \ourdataset, a multilingual benchmark comprising $928$ books generated by five recent LLMs across six languages and three scripts, with an average %
length of approximately $59$K words per book. The benchmark supports evaluation under domain, author, and language shifts. Evaluation of representative AA methods shows that no single method consistently performs best across settings, and performance generally degrades under distribution shifts. %
{\color{black}Transformer-based detectors can retain generator-related information across languages, although transfer effectiveness varies by language pair, whereas statistical and fingerprint-based detectors are more language-dependent. 
We envision \ourdataset as a valuable resource for the development and evaluation of robust AA methods.}\footnote{The dataset and code can be found at \url{https://github.com/GrecoMT/MultiGhostBench}.}

\end{abstract}

\section{Introduction}
Large language models (LLMs) can now generate high-quality text across multiple languages \citep{multilinguallargelanguagemodel,pangea}, achieving levels of fluency and coherence comparable to human-written content \citep{jakesch}. These advances create new opportunities for communication, creativity, and productivity \citep{mega,acm-survey-multiling}. Still, they also raise concerns about transparency, accountability, misuse, and the authenticity of digital content \citep{brundage2024malicioususeartificialintelligence,ChatGPT-academia,russell-etal-2026-ai}. These risks motivate the need for robust methods to detect and attribute LLM-generated text \citep{AAinLLMera,wu2025survey}.

Although prior research has explored LLM-generated text detection and numerous datasets have been introduced, several limitations remain as outlined in \reftab{tab:existing-datasets-overview}. First, most formulate detection as a binary human-versus-AI problem \citep{MGTBench, RAID,thai2026editlens}, while only a limited number of works study the more challenging authorship attribution (AA) task \citep{turingbench, OTB, ghostwritebench},
where the objective is to identify the specific generator LLM that wrote a text. Second, existing datasets predominantly focus on short-form texts \citep{multisocial,multitude}, despite the emergence of LLM-ghostwritten books \citep{ausocietyauthors}. Even recent long-form efforts, such as \citet{ghostwritebench}, remain monolingual or limited to a small number of languages \citep{RAID, CUDRT}. Third, methods are rarely evaluated for multiple distribution shifts, particularly cross-language transfer or generalisation to unseen generators and domains. 

In this paper, we introduce \ourdataset, a multilingual benchmark for LLM AA that evaluates generalisation under domain, generator, and language shifts. To the best of our knowledge, \ourdataset is the first benchmark designed to jointly evaluate multilingual long-form LLM AA under multiple distribution shifts. To summarise, our main contributions are:

\begin{itemize}
    \item We introduce \ourdataset, comprising $928$ books (average $59$K words) generated by five recent LLMs. It covers six languages and three scripts, and is designed to evaluate AA both within and across languages under domain and generator shifts.
    \item We conduct a comprehensive evaluation of statistical, supervised, and fingerprint-based attribution detectors, finding that no method performs consistently best across languages, data regimes, and distribution shifts.
    \item Our analysis reveals that Transformer-based detectors can retain generator-related information across languages, although the extent of this transfer varies across language pairs, whereas statistical and fingerprint-based approaches are more strongly affected by language.
\end{itemize}

\section{Related Work}

\paragraph{Authorship Attribution.}

Research on LLM-generated text has predominantly focused on binary detection \citep{mitchell2023detectgpt,yang2024dnagpt,thai2026editlens}, with only a few works \citep{turingbench,OTB,ghostwritebench} studying the AA task, where the goal is to identify the specific LLM that generated a text. However, these studies are predominantly based on supervised methods with limited generalisation. \citet{ghostwritebench} studied long-form texts and robustness under domain and unseen-generator shifts, but their study remained restricted to English. They also developed a lightweight fingerprint-based method, \trace, for the AA task.

\paragraph{Multilingual Authorship Attribution.}
Similarly, research on multilingual LLM-generated text has primarily focused on binary detection \citep{detectrl,ceaid,m4gt} rather than attribution. \textsc{MULTITuDE} \citep{multitude} evaluates cross-language attribution performance, but its evaluation is not exhaustive, as it does not cover all languages in the dataset. Likewise, attribution experiments in \textsc{M4GT-Bench} \citep{m4gt} focus on cross-domain rather than cross-lingual generalisation. More recently, \citet{la-cava-authorship} provide a systematic study of multilingual attribution using \textsc{MULTITuDE} and \textsc{MultiSocial}. However, their primary setting is based on short news articles. This misses practical scenarios involving unseen generators and joint evaluation across multiple dimensions as done in this work, which we summarise in \reftab{tab:existing-datasets-overview}.

\begin{table}[!htp]
    \centering
    \resizebox{\columnwidth}{!}{%
    \begin{tabular}{l @{\ }
                    S[table-format=2.0] @{\ }
                    c @{\ \ \ }
                    c @{\ \ \ }
                    c @{\ \ \ }
                    c @{\ \ \ }
                    S[table-format=2.0] @{\ \ \ \ }
                    c @{}
                    r @{\ }
                    r}
        \toprule[1.5pt]
        \multirow{2}{*}{\textbf{Dataset}}
        & \multicolumn{2}{c}{ \textbf{LLMs}}
        & \multicolumn{3}{c}{\textbf{OOD}}
        & \multicolumn{1}{c}{\multirow{2}{*}{\rotatebox{90}{{\small\# }\textbf{Lang}}}}
        & \multirow{2}{*}{\rotatebox{90}{\textbf{Long?}}}
        & \multicolumn{1}{c}{\textbf{Avg.}}
        & \multicolumn{1}{c}{\textbf{Num}}
        \\
        \cmidrule(lr){2-3}
        \cmidrule(lr){4-6}
        & \textbf{\#}
        & \textbf{Rec.}
        & \textbf{\small D}
        & \textbf{\small A}
        & \textbf{\small L}
        &
        &
        & \multicolumn{1}{c}{\textbf{Words}}
        & \multicolumn{1}{c}{\textbf{Docs}}
        \\
        \midrule
        \textsc{TuringBench}\textsubscript{(\citeyear{turingbench})}
        & 20
        & \xmark
        & \xmark
        & \xmark
        & \xmark
        & 1
        & \xmark
        & $<$$200$
        & $160$K
        \\
        \textsc{OpenTuring}\textsubscript{(\citeyear{OTB})}
        & 7
        & \xmark
        & \cmark
        & \cmark
        & \cmark
        & 1
        & \xmark
        & $<$$500$
        & $497$K
        \\
        {\textsc{GhostWriteB.}\textsubscript{(\citeyear{ghostwritebench})}}
        & 10
        & \cmark
        & \cmark
        & \cmark
        & \xmark
        & 1
        & \cmark
        & $53$K
        & $325$
        \\
        \textsc{M4GT-Bench}\textsubscript{(\citeyear{m4gt})}
        & 6
        & \xmark
        & \cmark
        & \xmark
        & \xmark
        & 12
        & \xmark
        & $<$$500$
        & $87$K
        \\
        \textsc{MULTITuDE}\textsubscript{(\citeyear{multitude})}
        & 8
        & \xmark
        & \xmark
        & \xmark
        & \cmark
        & 11
        & \xmark
        & $<$$512$
        & $74$K
        \\
        \textsc{MultiSocial}\textsubscript{(\citeyear{multisocial})}
        & 8
        & \xmark
        & \xmark
        & \xmark
        & \cmark
        & 22
        & \xmark
        & $<$$200$
        & $472$K
        \\
        \midrule
        \textbf{\ourdataset}
        & 5
        & \cmark
        & \cmark
        & \cmark
        & \cmark
        & 6
        & \cmark
        & \bm{$59$}\textbf{K}
        & \bm{$928$}
        \\
        \bottomrule[1.5pt]
    \end{tabular}%
    }
    \vspace{-0.5em}
    \caption{
    Overview of existing datasets for LLM-generated text attribution. `Rec.' indicates whether benchmark includes recent high-capability LLMs; prior benchmarks include small-scale models ($<$$10$B parameters). `D', `A', and `L' denote OOD-Domain, OOD-Author, and OOD-Language, respectively. `Long?' denotes whether documents longer than $10$K words.
    }
    \label{tab:existing-datasets-overview}
\end{table}

\section{Dataset}
\label{sec:dataset}

Our proposed \ourdataset is a multilingual dataset of long-form texts generated by five 
recent
LLMs (\geminiPro, \geminiFlash, \deepseek, \qwen, and \gptOSS; more details in \refapptab{tab:llm-model-card})
covering six languages and supporting AA under 
domain, author, and language shifts. A unique feature of our dataset is its %
language-shift setting, which enables the evaluation of whether attribution methods trained on one language can generalise to texts written in a previously unseen language. %

\subsection{Language Selection}
\label{subsec:language_selection}

\ourdataset covers six languages: Italian (\ita{}), Spanish (\es{}), German (\de{}), English (\en{}), Chinese (\zh{}), and Russian (\rus{}). These languages represent four language families: Romance (\ita{} and \es{}), Germanic (\de{} and \en{}), Sino-Tibetan (\zh{}), and Slavic (\rus{}). They also cover multiple writing systems, including Latin, Chinese characters, and Cyrillic scripts as well as language-specific orthographic conventions. This selection enables evaluation of AA under both closely related and more distant language and script shifts.

\subsection{Multilingual Book Generation Pipeline}
\label{subsec:generation_pipeline}

Human book writing is generally hierarchical: authors typically develop a book through multiple stages, such as planning, drafting, and revision \cite{sun-etal-2022-summarize, lee-etal-2025-navigating}. We follow the generation process introduced by \citet{ghostwritebench} that simulates this multi-stage writing process to create the LLM-generated books.

{\color{black}The book generation pipeline begins with the definition of a set of constraints,}
such as genres and time period. These constraints are sourced from Project Gutenberg\footnote{\url{https://www.gutenberg.org/}}, an online library of freely available books; see \refapptab{tab:genres-info} for the list of genres, comprising both fiction and non-fiction genres. 
A Gutenberg book is usually multi-genre (\eg \textsf{LF} (Literature \& Fiction) $+$ \textsf{HB} (History \& Biographies)), therefore books in \ourdataset are also multi-genre. 
Each book is  expanded iteratively by generating each subsequent segment conditioned on the outline, the previous segment, and a running summary of the narrative, \ie updated after each generation step. This structure allows models to preserve global coherence while producing texts that exceed the length of a single generation step. We adapt the original prompt templates from \citet{ghostwritebench} for the multilingual setting by translating them into each target language. The translations were verified by respective native-speaker volunteers; prompts can be found in \refapp{sec:gen-prompts}.

We provide model details for five LLMs and quantitative statistics for these LLM-generated books in \refapptab{tab:llm-model-card}~and~\ref{tab:quantitative_textual_metrics}, respectively. All the books were generated consistently using the same pipeline described above to minimise pipeline-specific bias. 
However, some LLM-specific generation bias is inevitable, which we try to reduce through our meticulous data cleaning (\S\ref{subsec:data_cleaning}) and analysis (\S\ref{subsec:data_exploration}).

\begin{table}[t!]%
\centering
\resizebox{0.9\columnwidth}{!}{%
\begin{tabular}{lS[table-format=2.0]S[table-format=2.0]S[table-format=2.0]S[table-format=2.0]S[table-format=2.1]}
\toprule[1.5pt]

\multirow{2}{*}{\rotatebox{0}{\textbf{Lang.}}} & \multicolumn{4}{c}{\textbf{\# Books}} & \textbf{Avg. \#} \\
\cmidrule(lr){2-5}
    & \multicolumn{1}{c}{Train} & {Dev.} & {Test} & {Total} & \textbf{Words} \\
\midrule
\en{} & 92 & 10 & 61 & 163 & 64.8K \\
\de{} & 88 & 10 & 56 & 154 & 44.4K \\
\ita{} & 93 & 10 & 58 & 161 & 50.2K \\
\es{} & 93 & 10 & 60 & 163 & 44.9K \\
\rus{} & 79 & 10 & 52 & 141 & 83.6K \\
\zh{} & 82 & 10 & 54 & 146 & 69.7K \\
\midrule
\textbf{Total} & 527 & 60 & 341 & 928 & 59.6K \\
\bottomrule[1.5pt]
\end{tabular}%
}
\vspace{-0.5em}
\caption{\ourdataset statistics: number of training, development, and test books, together with the average book length (in words) for each language. See \refapptab{tab:dataset_genre_splits} for detailed breakdown per LLM. }
\label{tab:dataset_summary}
\end{table}

\subsection{OOD Dimensions}
\label{subsec:ood_dimension}

Following \citet{ghostwritebench}, \ourdataset supports two OOD evaluation dimensions: \textbf{\oodOne}, which measures generalisation to unseen genres, and \textbf{\oodTwo}, which measures generalisation to unseen generators. We further introduce \textbf{\oodThree}, which evaluates whether AA methods transfer to texts written in languages not observed during training.

In \oodOne, genres are partitioned into ID and OOD subsets for each generator LLM, ensuring that no genre appears in both partitions. Recall that these genres are sourced from the Gutenberg corpus %
comprising both fiction and non-fiction genres. Full statistics for genres in ID and \oodOne for different LLMs are provided in \refapptab{tab:dataset_genre_splits}. In \oodTwo, we adopt a leave-one-author-out protocol: one LLM is held out at a time, methods are trained on the remaining LLMs, and evaluation is performed on books generated by the held-out model. In \oodThree, methods are trained on one source language and evaluated on a different target language. We do not combine \oodThree with \oodOne, so that the evaluation isolates the effect of language shift rather than conflating it with genre shift.

\subsection{Data Cleaning}
\label{subsec:data_cleaning}
We clean the generated texts to remove generation and encoding artefacts while preserving language-specific features. 
Invalid control characters, markup, and non-textual symbols are removed, while formatting elements are normalised, preserving language-specific punctuation, diacritics, and non-Latin scripts.
Following \citet{ghostwritebench}, we remove $1.5$K tokens from both the beginning and end of each book to reduce shortcut cues from identifier markers, such as author names, titles, or publication years. As a final sanity check, we use \texttt{GlotLID} \citep{glotlid}, an open-source language identification model, to verify that each generated book is written in its intended target language; all books in \ourdataset match their intended target language.

\subsection{Data Exploration}
\label{subsec:data_exploration}

\reftab{tab:dataset_summary} shows a summary statistics for \ourdataset, including   the number of books per language across different splits. To provide a consistent estimate of document length across languages, we tokenise each book using a language-specific tokeniser (listed in \refapptab{tab:language_tokenizers}) and convert the token counts to approximate word counts using the standard 1.5 tokens-per-word conversion.

\paragraph{Quantitative analysis.}
\label{sec:quant-analysis}
The generated books are analysed using multiple quantitative textual metrics in \refapptab{tab:quantitative_textual_metrics}. We compute perplexity (PPL) as a model-based measure of text predictability \citep{jm3}, n-gram-based metrics for lexical overlap and diversity (\selfBleu (S-B), self-repetition (Self-R), and $n$-gram diversity (NGD)) \citep{self-bleu,salkar-etal-2022-self,meister-etal-2023-locally}, and compression ratio as a proxy for textual redundancy \citep{shaib-etal-2025-standardizing}. Given the multilingual nature of \ourdataset, token-based metrics are computed using language-dependent tokenisers (\refapptab{tab:language_tokenizers}).

For comparison, we select human-written books from the Gutenberg corpus with similar properties (full-length, single-authored works from the nineteenth or twentieth century). Overall, across languages, the generated books exhibit textual properties comparable to human-written books, particularly in lexical diversity and redundancy. However, generated books show higher S-B scores and lower PPL, indicating greater inter-book lexical similarity and higher predictability for the reference model (\ie \mgpt). For Chinese and Russian, no eligible human references were available, hence they were compared  with LLM-generated books in other languages. Their metric values remain within a similar range, with some variation for Russian PPL and Chinese NGD.

\paragraph{Human evaluation.}

We conduct a small-scale human evaluation on a subset of generated books as an initial validation step. Our   focus is on novels generated by \geminiPro in the \textsf{LF} (Literature \& Fiction) genre. Following prior works \citep{chhun-etal-2022-human, wang-etal-2025-towards-novel, ghostwritebench}, the generated books are evaluated along several dimensions that capture both local linguistic quality and broader narrative structure (more in \refapp{sec:human-eval}).

\refapptab{tab:human_eval} summarises the annotator scores across passage-level and overall book-level dimensions. Overall, the results indicate that the generated books are generally fluent and coherent with strong passage-level scores, while narrative-level quality is less consistent across languages for subjective dimensions. Although this evaluation is limited to \geminiPro-generated literature books, the findings complement our   analyses above, which show similar quality trends across LLMs. Together, these results support   the overall quality of books in \ourdataset. %

\section{Authorship Attribution Methods}

The selected methods span three categories: \textit{metric-based}, \textit{model-based supervised}, and \textit{fingerprint-based}. Their multilingual adaptations are summarised below, with further detector and implementation details in \refapp{app:authorship-attribution-methods}~and~\ref{app:impl_det}, respectively.

\paragraph{Metric-based methods.} For this category, we consider \rank, \entropy, and \gltr~\citep{gehrmann-etal-2019-gltr}. These methods rely on token-level probabilities produced by a reference language model. Following \citet{multitude}, we use \mgpt~\citep{mgpt} as the multilingual reference model. We adapt their extracted statistics to multi-class attribution following \citet{OTB,ghostwritebench}.

\paragraph{Model-based supervised methods.} We consider \ngram~\citep{koppel2004authorship}, \bert~\citep{fabien-etal-2020-bertaa}, and \detective~\citep{guo2024detective}, covering both non-neural and neural detectors. For \ngram, we adapt preprocessing to preserve accented and language-specific characters and use \textit{Jieba}\footnote{\url{https://pypi.org/project/jieba/}} segmentation for Chinese word-level features. For \bert, we use \xlmroberta~\citep{conneau-etal-2020-unsupervised} as the multilingual encoder. For \detective, we replace the original sentence-embedding model with a multilingual counterpart.

\paragraph{Fingerprint-based methods.} We consider \trace~\citep{ghostwritebench}, a post-hoc fingerprint-based AA method that models transitions between token-rank or entropy statistics computed by an evaluator language model. A test text is attributed to the generator with the most similar fingerprint. We evaluate \tracerank, \traceentjs, and \traceentnor. For multilingual evaluation, we replace the original evaluator, \gpt, with \gemma, also evaluated by \citet{ghostwritebench}. Previous \trace evaluations were limited to English.

\begin{figure*}[t]
    \centering
    \includegraphics[
        width=0.95\textwidth,
        trim=0 0 0 0,
        clip
    ]{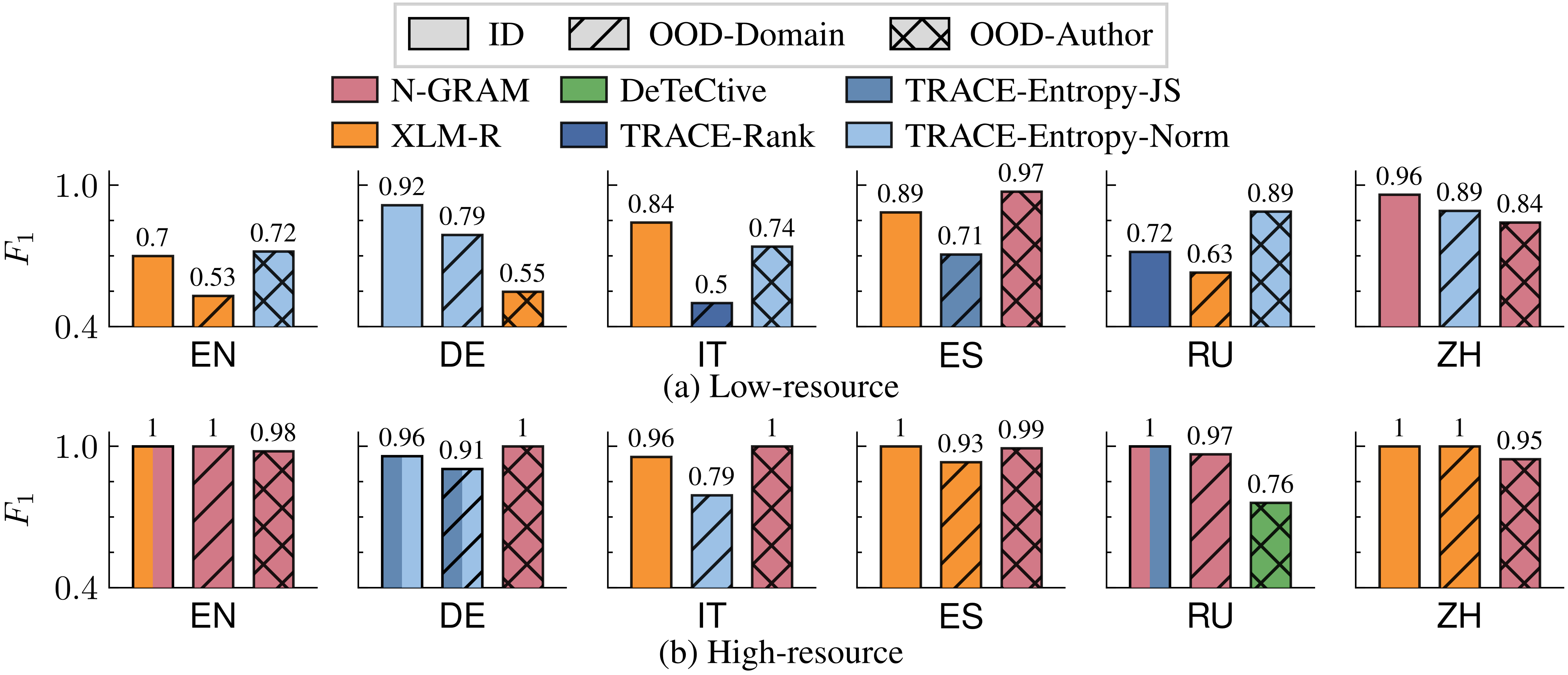}
    \vspace{-0.75em}
    \caption{
    Best-performing detector by language and evaluation setting under the
    low- and high-resource regimes.
    Each group includes ID, \oodOne, and \oodTwo.
    Bar height shows macro-$F_1$, colors identify detectors, hatches indicate
    evaluation settings, and split colors denote ties.
    The best detector varies across conditions, while more data
    generally improves performance.
    }

    \label{fig:best-detectors-by-language}
\end{figure*}

\section{Experimental Setup}

\paragraph{Training settings.}
In order to analyse how AA performance varies with respect to the amount of available training data, we follow \citet{ghostwritebench} considering two training scenarios: \textbf{high-resource} setting where a relatively large number of labelled training books is available for each LLM (10--30 books), and \textbf{low-resource} setting where only a limited number of labelled training books is available for each LLM (1--5 books).

\paragraph{Evaluation settings.}
Since \ourdataset includes the \oodTwo dimension, the generator of a test text may not appear among those observed during training. This corresponds to an open-set setting, where an AA method must not only distinguish among known authors, but also reject texts generated by unseen authors. 
To support this rejection mechanism, all methods are calibrated on the development set 
by selecting a confidence threshold that best balances attribution performance and rejection ability. All the thresholds can be found in \refapptab{tab:thresholds}.

During evaluation, the same threshold is applied across all scenarios: ID, \oodOne, \oodTwo, and \oodThree. We do this because in a practical application we would not know whether a test document is in or out of the training distribution.
For texts generated by known authors, predictions below the confidence threshold are counted as missed attributions. For texts generated by unseen authors, the same behaviour is instead counted as a correct rejection, since the method avoids forcing the sample into one of the known author classes. Attribution performance is evaluated using macro-F1, which gives equal weight to all generating LLMs.

\section{Experimental Results}
\vspace{-1mm}
We first consider in-language evaluation (\S\ref{sec:in-lang-results}), where training and test texts are written in the same language, covering the ID, \oodOne, and \oodTwo settings.
We then analyse cross-language evaluation (\S\ref{sec:cross-lang-results}), where attribution methods are trained on one source language and evaluated on a previously unseen target language.
Finally, we examine the learned representation (\S\ref{sec:rep-analysis}) space to better interpret the observed performance patterns under domain and language shifts.

\subsection{In-Language Evaluation}
\label{sec:in-lang-results}

\reffig{fig:best-detectors-by-language} reports the best-performing method for each
language and evaluation setting; complete results are reported in \refapptab{tab:results_low}~and~\ref{tab:in-lang-results-high}. For completeness, the results obtained without confidence-threshold rejection are provided in \refapptab{tab:results_low_without_threshold}~and~\ref{tab:results_high_without_threshold}.

\paragraph{No universal winner.}
No single method consistently outperforms all others across languages, resource regimes, and evaluation settings. Nevertheless, some methods appear more frequently among the best-performing approaches. \xlmroberta is
particularly competitive in the ID and \oodOne settings, while \ngram becomes a frequent winner in the high-resource regime, especially under \oodTwo evaluation. \trace variants also obtain the best result in several language-specific configurations. In contrast, \rank, \entropy, and \gltr consistently achieve low performance, indicating that individual token-level statistics provide insufficient signals for reliable authorship attribution.

\begin{figure*}[t]
    \centering
    \includegraphics[
        width=0.9\linewidth
    ]{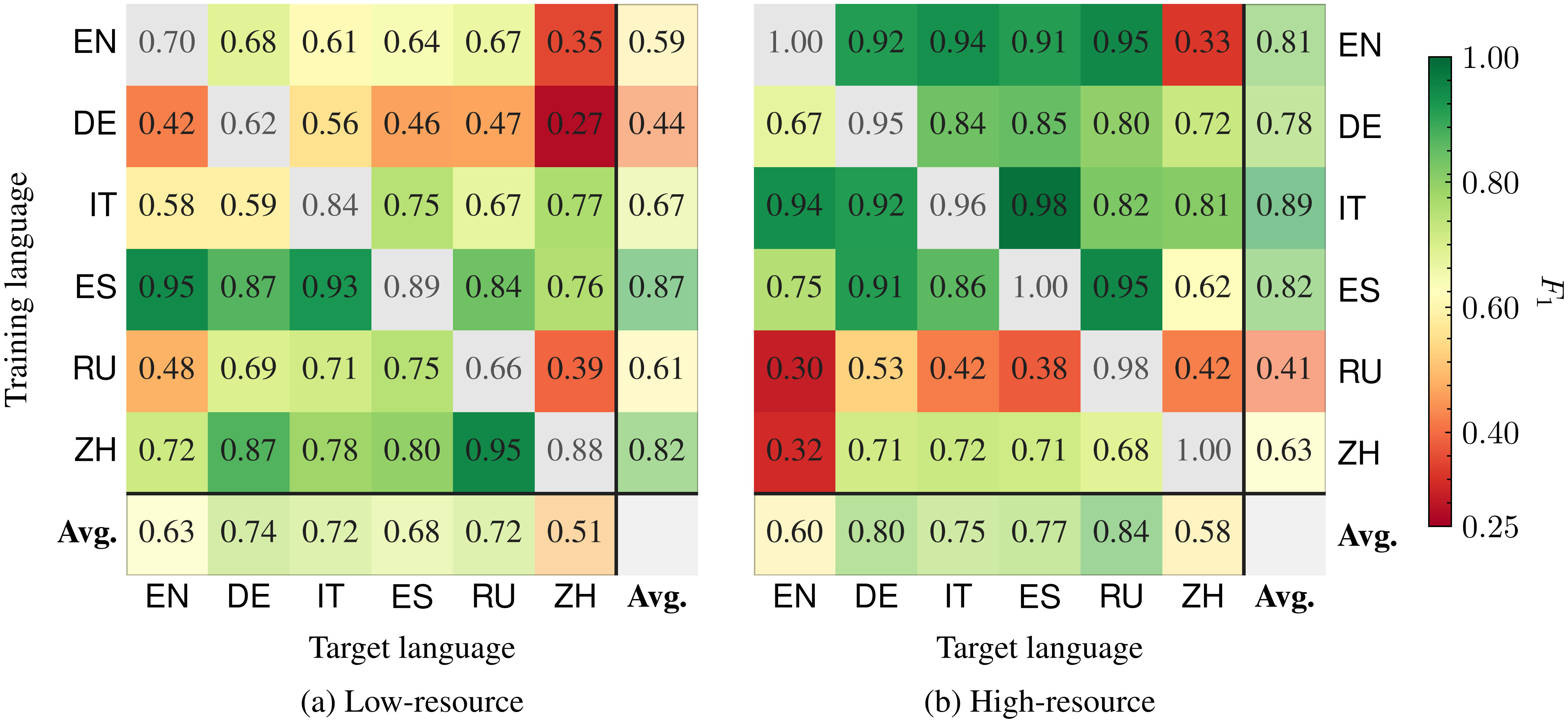}
    \vspace{-0.75em}
    \caption{
    Cross-language macro-$F_1$ results for \xlmroberta in the
    (a) low-resource and (b) high-resource settings.
    Rows indicate source languages and columns target languages;
    the final row and column report averages excluding the diagonal.
    Transfer depends on both the source-target pair and the resource regime; Chinese remains the most challenging target.
    }
    \label{fig:xlmr-cross-language-heatmaps}
\end{figure*}

\paragraph{More data improves performance.}

In the high-resource setting, the larger amount of training data improves the best achievable performance across all ID and \oodOne configurations. In several cases, the best-performing detector approaches or reaches perfect macro-$F_1$. 

Full results (Tables~\refapptab{tab:in-lang-results-high} and~\ref{tab:results_low}) show the impact of additional training data on each detector. Under \oodTwo, \xlmroberta performs worse than in the low-resource setting for Italian, Chinese, and English, whereas \detective improves across all languages. \ngram benefits the most from the additional data, becoming the best-performing \oodTwo detector in five languages and the second-best in Russian.
An explanation is that \ngram needs more data to estimate reliable lexical and structural patterns, whereas \xlmroberta may become overconfident on texts from unseen authors, assigning them to one of the known classes instead of rejecting them.

\paragraph{Robustness drops under OOD shifts.}
Compared with ID evaluation, performance generally decreases under distribution shift, although the strongest high-resource detectors retain high absolute scores. The OOD-ID differences reported in \refapptab{tab:low_resource_deltas}~and~\ref{tab:high_resource_deltas} are negative in most cases, with \oodTwo generally producing larger drops than \oodOne.

\paragraph{Language differences.}
The low-resource setting reveals clearer performance differences across languages. Chinese achieves the strongest results, whereas English is comparatively more challenging. Indeed, %
the generators produce Chinese texts of more variable quality, creating stronger generator-specific signals that attribution detectors can exploit; in contrast, English outputs' quality may be consistently better, leading to greater stylistic overlap and weaker attribution cues. %
These differences become less pronounced in the high-resource regime, where most best-performing scores approach or reach perfect macro-$F_1$.

\subsection{Cross-Language Evaluation}
\label{sec:cross-lang-results}

\reffig{fig:xlmr-cross-language-heatmaps} reports the cross-language macro-F1 scores of \xlmroberta under the low- and high-resource settings. Rows indicate the source language used for training, while columns indicate the target language used for evaluation. 
We focus the visual analysis on \xlmroberta because it achieves the strongest overall cross-language performance. \detective, the other Transformer-based method, also transfers substantially better than the remaining approaches, but generally obtains lower scores than \xlmroberta, despite outperforming it in a limited number of configurations. By contrast, the metric-based, N-GRAM, and fingerprint-based methods achieve near-zero performance in most cross-language configurations, suggesting that the signals they capture are largely language-dependent and do not transfer reliably across languages. For completeness, the full results for all methods with and without thresholding are reported in \refapptab{tab:ood_language_threshold_all} and \reftab{tab:ood_language_no_threshold_all}, respectively.

\begin{figure*}[t!]
    \centering
    \includegraphics[
        width=0.9\textwidth,
        trim={0 0 0 0},
        clip
    ]{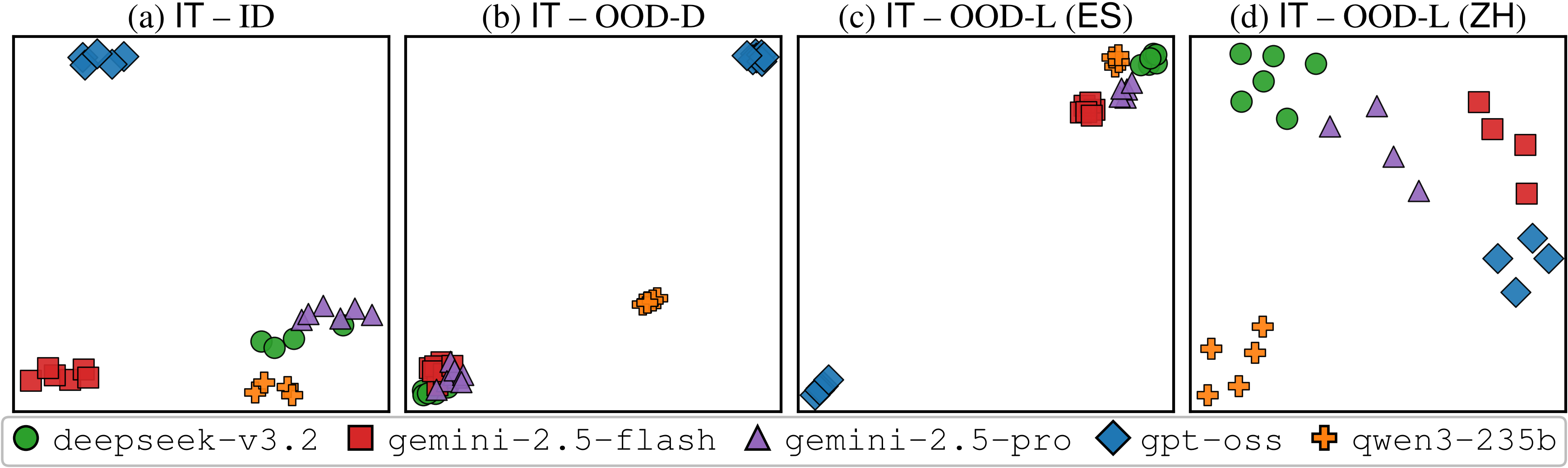}
\caption{UMAP projections of \xlmroberta representations after fine-tuning on Italian.
    The panels show evaluation on (a) Italian in-distribution (ID), (b) Italian
    \oodOne, and (c) Spanish and (d) Chinese ID texts evaluated under \oodThree.
    Generator-specific clusters remain visible in all settings, with clearer
    separation for Spanish than under domain shift or for Chinese.
}
    \label{fig:xlmr-italian-four-panels}
\end{figure*}

\paragraph{Target language difficulty.} 
Chinese is the most challenging target language. Averaging across source languages, transfer to Chinese achieves a macro-F1 of 0.508 in the low-resource setting and 0.582 in the high-resource setting, the lowest average in both regimes. Russian provides an informative contrast. Although it also uses a non-Latin script, it obtains an average target score of 0.717 in the low-resource setting and 0.841 in the high-resource setting, making it the easiest target language in the latter regime. Therefore, the difficulty of transferring to Chinese cannot be explained by script differences alone. Broader typological differences, tokenisation behaviour, and language-specific distributional properties may also contribute to the observed performance gap. In particular, the distinctive NGD values observed for Chinese in the quantitative analysis may reflect distributional characteristics that make cross-language transfer more difficult, although these descriptive metrics do not directly establish the source of the performance degradation.

\paragraph{Source language effects.}

The effectiveness of a source language varies substantially across resource regimes. In the low-resource setting, Spanish provides the strongest overall transfer, with an average macro-F1 of 0.866 across target languages, followed by Chinese with 0.823. German is the weakest source language, with an average of 0.435. This ranking changes in the high-resource setting: Italian becomes the strongest source language, with an average macro-F1 of 0.894, followed by Spanish with 0.819, whereas Russian becomes the weakest at 0.410. Chinese also drops markedly, from 0.823 to 0.629, becoming the second-weakest source language in the high-resource regime. 
In particular, the deterioration observed for Chinese and Russian suggests that, in these languages, training with more data may lead to overfitting to source-language-specific patterns, reducing the transferability of the learned authorship signals to other languages.
The language-specific differences observed in Chinese NGD and Russian PPL may also reflect underlying distributional properties that contribute to this behaviour, although these metrics do not directly explain the transfer degradation.

\paragraph{Language family effects.}
The clearest evidence of a language-family effect concerns Italian and Spanish. In the low-resource setting, Spanish provides the best transfer to Italian, with a macro-F1 of 0.926, while Italian is the second-best source for Spanish, reaching 0.750.
With more training data, the relationship becomes fully reciprocal: Spanish-to-Italian reaches 0.857, while Italian-to-Spanish achieves 0.981, the highest score in the table. The Germanic pair (English and German) shows a weaker and less symmetric pattern. English is only the fourth-best source language for German in the low-resource setting, but becomes the strongest source in the high-resource setting, reaching a macro-F1 of 0.916. Conversely, German-to-English achieves only 0.420 in the low resource regime, but improves to 0.673 with more training data, becoming the third-best source language. Overall, language-family relatedness appears to favour transfer, particularly for the Romance pair, but it does not fully determine cross-language performance.

\subsection{Representation Analysis}
\label{sec:rep-analysis}
To better understand the behaviour of \xlmroberta and \trace beyond their attribution scores, we inspect their respective representation spaces. We use UMAP \cite{umap} to visualise how \xlmroberta embeddings and \trace fingerprints change across the ID, \oodOne, and \oodThree settings. The two methods offer complementary perspectives on the attribution task: \xlmroberta relies on learned representations, whereas \trace operates in a statistical fingerprint space. Comparing these spaces allows us to examine how the two approaches are affected by distribution shifts.

\paragraph{\xlmroberta embeddings.}
We focus on the representation space learned by \xlmroberta in the high-resource setting, using the model fine-tuned on Italian, which achieves the strongest average cross-language performance. We consider embeddings from Italian ID texts, Italian \oodOne texts, and Spanish and Chinese texts evaluated in the \oodThree setting to examine the effects of domain and language shifts. \reffig{fig:xlmr-italian-four-panels} illustrates   the embedding space produced by \xlmroberta   across the four evaluation settings. In the ID setting, texts generated by the same model form compact, well-separated clusters. Under \oodOne, this organisation is only partially preserved, with clusters becoming less compact and exhibiting greater overlap. This pattern is consistent with the decrease in macro-F1 from 0.956 in the ID setting to 0.676 under \oodOne, indicating that domain shift makes the learned representations less discriminative. 
When the Italian-fine-tuned model is evaluated on Spanish texts, the model-specific structure is largely preserved, with compact and well-separated clusters similar to those observed in the Italian ID setting. This is consistent with the strong Italian-to-Spanish transfer performance and suggests that the learned generator-related representations generalise effectively between closely related languages. In contrast, when the same model is evaluated on Chinese texts, a model-specific organisation remains visible, but the clusters are broader and less clearly separated than those observed for Spanish. This suggests that the learned representations retain generator-related information across languages, although their discriminative structure is increasingly affected by larger typological and script differences.

\paragraph{\trace fingerprints.}
We next analyse the entropy-based fingerprint space of \trace in the high-resource setting, using Italian as the reference language. While \trace remains comparatively stable in the same-language settings, its performance drops sharply under \oodThree evaluation. Since attribution is performed by comparing distances between test and training fingerprints, their relative position provides insight into this behaviour. \reffig{fig:trace_reference_comparison} compares Italian training, ID-test, and \oodOne fingerprints with fingerprints from different target languages. Italian ID-test and \oodOne samples remain close to the training distribution, whereas fingerprints from other languages occupy clearly displaced regions. Notably, 
this pattern is observed not only for Chinese, which is linguistically distant from Italian, but also for Spanish, a closely related Romance language. Consequently, cross-language samples lack nearby reference fingerprints, making distance-based attribution less reliable. This suggests that the fingerprint space is strongly influenced by language-specific token transition patterns, which helps explain the poor cross-language performance of \trace.

\section{Conclusion}
We introduced \ourdataset, a multilingual benchmark for LLM AA comprising 928 long-form books generated by five recent LLMs, and   covering six languages from four language families and three scripts. Importantly, it is designed to evaluate generalisation under domain, generator, and language shifts. For the within-language setting (training and testing on the same language), no single method consistently outperforms the others across languages, data regimes, or the ID, \oodOne, and \oodTwo conditions. In the cross-language setting (\oodThree), however, Transformer-based approaches (\xlmroberta and \detective) substantially outperform other methods. Interestingly, \trace and metric-based methods drop to near-zero performance. Further analysis shows that cross-language performance is influenced by source and target language, language-family proximity, and script differences. These results highlight limitations of current methods in multilingual settings. We hope \ourdataset will serve as a useful benchmark for developing robust attribution methods.

\section*{Limitations}

Although \ourdataset covers six languages spanning four language families and three writing scripts, it does not capture the full diversity of multilingual LLM-generated text. In particular, it does not cover low-resource languages, languages underrepresented in LLM pre-training data, or languages with substantially different grammatical and morphological characteristics. Furthermore, we consider five recent LLMs capable of generating long-form content across the selected languages. Extending the benchmark to additional and future LLMs is an important direction for future work. Our human evaluation is also limited to a small subset of generated books. Moreover, although the generated books are generally of good quality, multilingual long-form generation remains challenging, particularly with respect to originality, self-evaluation, revision, \etc across languages. These aspects are beyond the scope of this work. Finally, our study focuses on attribution under single-author and non-adversarial settings. Extensions such as mixed authorship (\ie a book written by multiple authors) and adversarial generalisation (such as obfuscation attacks) remain important directions for future research.

\section*{Ethical Considerations}
\ourdataset consists of LLM-generated books for multiple languages. As with any creative text, some books may contain writing or plot elements that reflect undesirable values. However, all LLMs used in this work incorporate built-in safety guardrails, reducing the likelihood of generating such content. This is further supported by our human evaluation, where no such concerning content was identified. Overall, we aim to adhere to the ACL Code of Ethics.\footnote{\url{https://www.aclweb.org/portal/content/acl-code-ethics}}

\section*{Acknowledgements}
This research was conducted during M.\ Greco's visit at the University of Melbourne, funded by the Erasmus+ Programme. 
We thank the volunteer annotators from the University of Melbourne and the University of Calabria for their contribution to the human evaluation. This research was supported by The University of Melbourne's Research Computing Services and the Petascale Campus Initiative. 
A.\ Shetty was supported by the Commonwealth through an Australian Government Research Training Program Scholarship (DOI: \url{https://doi.org/10.82133/C42F-K220}), Computing and Information Systems PhD Scholarship, and the Avashya Foundation.
A. Tagarelli was partly supported by the Horizon Europe project AI-CODE, GA No. 101135437. J.H.\ Lau was supported by
the Australian Research Council under Grant LP210200917 and DP240101006.

\bibliography{custom}

\clearpage
\appendix
\section*{Appendix}

\section{\ourdataset Generation}

\begin{table}[H]
\centering
\begin{tabular}{ll}
\toprule[1.5pt]
\textbf{Tag} & \textbf{Genre} \\
\midrule
\textsf{ACM} & Art, Culture \& Media \\
\textsf{BLP} & Business, Law \& Politics \\
\textsf{HB} & History \& Biographies \\
\textsf{JR} & Journals \& Reports \\
\textsf{LF} & Literature \& Fiction \\
\textsf{LHH} & Lifestyle, Health \& Hobbies \\
\textsf{SSP} & Social Sciences \& Philosophy \\
\textsf{ST} & Science \& Technology \\
\bottomrule[1.5pt]
\end{tabular}
\caption{Different genres used in this work, sourced from Project Gutenberg \citep{momen-etal-2025-filling}.}
\label{tab:genres-info}
\end{table}

\begin{table*}[!htp]
    \centering
    \begin{tabular}{lS[table-format=3.1]ll}
    \toprule[1.5pt]
    \textbf{Model Name} & \textbf{Context} & \textbf{Type} & \textbf{Checkpoint} \\ 
    \midrule
    \deepseek~{\footnotesize\citep{deepseek-v3}} & {163.8K} & {\scalebox{0.9}{\faLockOpen}\ Open} & {\footnotesize\texttt{deepseek/\deepseek}} \\
    \gptOSS~{\footnotesize\citep{gpt-oss}} & {128.0K} & {\scalebox{0.9}{\faLockOpen}\ Open} & {\footnotesize\texttt{openai/gpt-oss-120b}} \\
    \qwen~{\footnotesize\citep{qwen-3}} & {262.1K} & {\scalebox{0.9}{\faLockOpen}\ Open} & {\footnotesize\texttt{qwen/qwen3-235b-a22b-2507}} \\
    \midrule
    \geminiFlash~{\footnotesize\citep{gemini-2.5}} & {1.05M} & {\scalebox{0.9}{\faLock}\ Closed} & {\footnotesize\texttt{google/\geminiFlash}} \\
    \geminiPro~{\footnotesize\citep{gemini-2.5}} & {1.05M} & {\scalebox{0.9}{\faLock}\ Closed} & {\footnotesize\texttt{google/\geminiPro}} \\
    \bottomrule[1.5pt]
    \end{tabular}
    \caption{LLMs checkpoints used in \ourdataset. `Context' is the total input context supported. `Type' indicates whether model weights are public ({\scalebox{0.9}{\faLockOpen}\ Open}) or if it is proprietary LLM ({\scalebox{0.9}{\faLock}\ Closed}).
    }
    \label{tab:llm-model-card}
\end{table*}

\begin{table*}[t]
\centering
\setlength{\tabcolsep}{7pt}
\begin{tabular}{lllrrrrrr}
\toprule[1.5pt]
\textbf{Model} & \textbf{Split} & \textbf{Genres} 
& \textbf{\ita{}} & \textbf{\zh{}} & \textbf{\es{}} & \textbf{\de{}} & \textbf{\en{}} & \textbf{\rus{}} \\
\midrule

\multirow{3}{*}{\geminiPro}
& ID         & \textsf{LHH}, \textsf{HB} & 18 & 8  & 10 & 18 & 6  & 11 \\
& ID         & \textsf{HB}      & 11 & 9  & 12 & 6  & 11 & 10 \\
& \oodOne & \textsf{LF}     & 8  & 6  & 6  & 7  & 5  & 6 \\

\midrule

\multirow{2}{*}{\geminiFlash}
& ID         & \textsf{HB}, \textsf{ACM} & 30 & 16 & 30 & 25 & 31 & 19 \\
& \oodOne & \textsf{LHH}, \textsf{LF} & 8  & 5  & 8  & 7  & 6  & 6 \\

\midrule

\multirow{2}{*}{\deepseek}
& ID         & \textsf{LF}      & 19 & 29 & 27 & 23 & 28 & 23 \\
& \oodOne & \textsf{HB}     & 5  & 8  & 7  & 6  & 8  & 7 \\

\midrule

\multirow{5}{*}{\qwen}
& ID         & \textsf{SSP}, \textsf{LF}       & 8  & 13 & 8  & 18 & 10 & 8 \\
& ID         & \textsf{LF}            & 9  & 12 & 11 & 6  & 10 & 11 \\
& \oodOne & \textsf{BLP}, \textsf{HB}       & 2  & 1  & 1  & 2  & 4  & 2 \\
& \oodOne & \textsf{HB}            & 1  & 2  & 2  & 1  & 3  & 2 \\
& \oodOne & \textsf{BLP}, \textsf{LHH}, \textsf{HB}  & 1  & 4  & 3  & 4  & 1  & 2 \\

\midrule

\multirow{2}{*}{\gptOSS}
& ID         & \textsf{HB}, \textsf{LF}        & 24 & 18 & 22 & 16 & 24 & 18 \\
& \oodOne & \textsf{BLP}, \textsf{HB}, \textsf{SSP} & 7  & 5  & 6  & 5  & 6  & 5 \\

\bottomrule[1.5pt]
\end{tabular}%
\caption{Detailed ID and \oodOne genre splits by generating model, with the number of books reported for each language. Development-set books are
excluded from this breakdown.
}
\label{tab:dataset_genre_splits}
\end{table*}

\subsection{Book Generation Prompts}
\label{sec:gen-prompts}
For brevity, they are not listed here; they can be found in \url{https://github.com/GrecoMT/MultiGhostBench/blob/main/prompts.py}.

\subsection{Quantitative Analysis}

\begin{table}[t]
\centering
\resizebox{0.99\columnwidth}{!}{%
\begin{tabular}{@{}lll@{}}
\toprule[1.5pt]
\textbf{Lang.} & \textbf{Tokeniser} & \textbf{Reference} \\
\midrule
\ita{} & \textsc{Minerva-7B}   & {\footnotesize\cite{orlando-etal-2024-minerva}} \\
\de{}  & \textsc{GBERT}        & {\footnotesize\cite{chan-etal-2020-germans}} \\
\es{} & \textsc{BETO}         & {\footnotesize\cite{canete2023spanishpretrainedbertmodel}} \\
\zh{} & \textsc{Chinese-BERT} & {\footnotesize\cite{devlin2019bertpretrainingdeepbidirectional}} \\
\en{} & \textsc{GPT-2}        & {\footnotesize\cite{hf_canonical_model_maintainers_2022}} \\
\rus{} & \textsc{Russian-BERT} & {\footnotesize\cite{kuratov2019adaptationdeepbidirectionalmultilingual}} \\
\bottomrule[1.5pt]
\end{tabular}
}

\caption{Language-dependent tokenisers used for token-based quantitative textual metrics.}
\label{tab:language_tokenizers}
\end{table}

\begin{table*}[h]
\centering
\begin{tabular}{llccccc}
\toprule[1.5pt]
\textbf{Lang} & \textbf{Author} &
\textbf{PPL} $\downarrow$ & \textbf{S-B} $\downarrow$ & \textbf{Self-R} $\downarrow$ & \textbf{NGD} $\uparrow$ & \textbf{CR} $\downarrow$ \\
\midrule

\multirow{6}{*}{\ita{}}
& \geminiPro        & 17.48 & 0.016 & 11.36  & 1.38 & 2.75 \\
& \geminiFlash      & 14.99 & 0.017 & 10.12  & 1.58 & 2.86 \\
& \deepseek         & 18.31 & 0.027 & 10.59  & 1.48 & 3.14 \\
& \qwen             & 18.97 & 0.019 & 11.26  & 1.37 & 2.75  \\
& \gptOSS           & 15.59 & 0.019 & 11.32  & 1.33 & 2.92  \\
\cmidrule(lr){2-7}
& Human             & 30.07 & 0.004 & 11.08 & 1.484 & 2.56 \\

\midrule

\multirow{6}{*}{\es{}}
& \geminiPro        & 19.76 & 0.019 & 10.47 & 1.55 & 2.72 \\
& \geminiFlash      & 17.00 & 0.014 &  9.43 & 1.65 & 2.85 \\
& \deepseek         & 20.89 & 0.013 & 11.44 & 1.39 & 2.81 \\
& \qwen             & 20.29 & 0.019 &  9.44 & 1.73 & 2.71 \\
& \gptOSS           & 16.67 & 0.011 & 11.29 & 1.21 & 3.02 \\
\cmidrule(lr){2-7}
& Human             & 30.01 & 0.004 & 11.04 & 1.39 & 2.73 \\
\midrule

\multirow{6}{*}{\de{}}
& \geminiPro        & 20.12 & 0.021 & 10.19 & 1.67 & 2.70 \\
& \geminiFlash      & 16.52 & 0.020 &  9.51 & 1.73 & 2.88 \\
& \deepseek         & 21.49 & 0.017 & 11.53 & 1.38 & 2.96 \\
& \qwen             & 23.87 & 0.014 & 10.33 & 1.58 & 2.74 \\
& \gptOSS           & 18.47 & 0.029 & 11.04 & 1.48 & 2.89 \\
\cmidrule(lr){2-7}
& Human             & 39.62 & 0.004 &  9.33 & 1.76 & 2.68  \\
\midrule

\multirow{6}{*}{\en{}}
& \geminiPro        & 30.67 & 0.024 & 10.68 & 1.59 & 2.68 \\
& \geminiFlash      & 27.81 & 0.016 &  9.92 & 1.65 & 2.82 \\
& \deepseek         & 31.43 & 0.016 & 11.80 & 1.40 & 2.75 \\
& \qwen             & 28.74 & 0.013 & 11.63 & 1.29 & 2.68 \\
& \gptOSS           & 28.84 & 0.016 & 11.44 & 1.39 & 2.79 \\
\cmidrule(lr){2-7}
& Human             & 31.36 & 0.001 & 10.25 & 2.14 & 2.65 \\
\midrule

\multirow{5}{*}{\rus{}}
& \geminiPro        & 15.46 & 0.020 & 11.47 & 1.61 & 3.66 \\
& \geminiFlash      & 12.16 & 0.022 &  9.58 & 1.90 & 3.98 \\
& \deepseek         & 15.66 & 0.016 & 10.21 & 1.56 & 3.69 \\
& \qwen             & 16.07 & 0.022 & 11.78 & 1.44 & 3.96 \\
& \gptOSS           & 16.98 & 0.022 & 11.88 & 1.63 & 3.63 \\

\midrule

\multirow{5}{*}{\zh{}}
& \geminiPro        & 19.19 & 0.027 & 11.92 & 1.10 & 2.27 \\
& \geminiFlash      & 14.16 & 0.021 & 10.54 & 1.20 & 2.50 \\
& \deepseek         & 24.55 & 0.019 & 12.27 & 1.08 & 2.29 \\
& \qwen             & 32.17 & 0.010 & 11.60 & 1.09 & 2.33 \\
& \gptOSS           & 21.40 & 0.022 & 11.64 & 1.00 & 2.56 \\

\bottomrule[1.5pt]
\end{tabular}
\caption{Average quantitative textual metrics for human and LLM-generated books across languages. The human reference consists of 100 books randomly sampled per language from the Project Gutenberg corpus released by \citet{momen-etal-2025-filling}, retaining only full-length, single-authored works originally written in the target language and published during the nineteenth or twentieth century, while excluding translations, short texts, and multi-authored works. Human values are reported only for languages for which a suitable reference sample was available.
}
\label{tab:quantitative_textual_metrics}
\end{table*}

We evaluate the generated books using several reference-free metrics capturing lexical diversity, repetition, and redundancy. Given the multilingual setting, token-based metrics are computed using language-specific tokenisers to reduce distortions caused by differences in tokenisation across languages. The analysis includes the following metrics:

\begin{itemize}
    \item \textbf{Self-BLEU (S-B):} measures the lexical similarity between books generated by the same model by computing BLEU scores between each book and the remaining outputs \citep{self-bleu}. Higher values indicate greater overlap and, consequently, lower inter-book diversity.

    \item \textbf{N-gram Diversity (NGD):} measures the variety of lexical patterns in the generated texts based on the proportion of distinct $n$-grams \citep{meister-etal-2023-locally}. We consider $n$-grams of length four, with higher values indicating greater diversity.

    \item \textbf{Self-Repetition (Self-R):} measures the extent to which $n$-grams are repeated within the generated texts, capturing the tendency of a model to reuse the same lexical patterns \citep{salkar-etal-2022-self}. Lower values indicate less repetition.

    \item \textbf{Compression Ratio (CR):} measures text redundancy through gzip compression \citep{shaib-etal-2025-standardizing}. More repetitive and predictable texts are generally more compressible, making this metric complementary to $n$-gram-based measures.

    \item \textbf{Perplexity (PPL):} measures how predictable a text is according to a reference language model. We compute perplexity using \mgpt \citep{mgpt}, providing a shared multilingual evaluation model across languages. Lower values indicate that the text is more predictable for the reference model.
\end{itemize}

\subsection{Human Evaluation}
\label{sec:human-eval}

\begin{table*}[t]
\centering
\begin{tabular}{lccccc|ccccc}
\toprule[1.5pt]
\multirow{2}{*}{\textbf{Lang.}} 
& \multicolumn{5}{c|}{\textbf{Passage-level}} 
& \multicolumn{5}{c}{\textbf{Overall}} \\
\cmidrule(lr){2-6} \cmidrule(lr){7-11}
& Rel. & Eng. & Coh. & Flu. & Div. 
& Coh. & Emp. & Sur. & Eng. & Comp. \\
\midrule
\en{} & 4.8 & 4.4 & 4.9 & 4.6 & 4.7 & 4.5 & 3.5 & 4.0 & 4.0 & 3.5  \\
\de{} & 4.9 & 3.7 & 4.4 & 4.0 & 3.6 & 5.0 & 4.5 & 3.0 & 3.0 & 3.5  \\
\ita{} & 4.8 & 4.1 & 4.6 & 4.2 & 4.8 & 5.0 & 5.0 & 3.5 & 4.0 & 4.0  \\
\es{} & 3.7 & 3.3 & 4.7 & 4.5 & 4.6 & 3.5 & 4.0 & 3.5 & 3.5 & 4.0  \\
\rus{} & 4.9 & 2.6 & 3.6 & 3.0   & 2.8 & 3.5 & 1.5 & 2.5 & 1.5 & 2.0    \\ 
\zh{} & 4.9 & 4.3 & 4.5 & 4.6 & 4.6 & 3.5 & 4.5 & 2.5 & 3.5 & 3.5  \\
\midrule
\textbf{Overall} 
& \textbf{4.66} & \textbf{3.73} & \textbf{4.45} & \textbf{4.15} & \textbf{4.18} 
& \textbf{4.16} & \textbf{3.83} & \textbf{3.16} & \textbf{3.25} & \textbf{3.41} \\
\bottomrule[1.5pt]
\end{tabular}%
\caption{Human evaluation scores across passage-level and overall narrative quality dimensions for \geminiPro generated books in the Literature \& Fiction (\textsf{LF}) genre.}
\label{tab:human_eval}
\end{table*}

Following \citet{ghostwritebench, chhun-etal-2022-human, wang-etal-2025-towards-novel}, we evaluate the generated books along several dimensions capturing linguistic, narrative, and structural properties at the micro-, meso-, and macro-levels:

\begin{itemize}
    \item \textbf{Relevance (Rel.):} measures the degree to which the generated text follows the given instructions and remains consistent with the specified book outline.

    \item \textbf{Engagement (Eng.):} measures the extent to which the text sustains the reader's interest and attention.

    \item \textbf{Coherence (Coh.):} evaluates how logically the narrative develops, considering the consistency of events, characters, and settings, as well as the progression from beginning to end.

    \item \textbf{Fluency (Flu.):} measures the grammatical correctness, readability, and naturalness of the language.

    \item \textbf{Diversity (Div.):} measures the degree of variation in vocabulary, sentence structure, and linguistic expression.

    \item \textbf{Empathy (Emp.):} evaluates how effectively the text conveys the characters' emotions, motivations, and reactions.

    \item \textbf{Surprise (Sur.):} measures the degree to which the ending introduces unexpected or non-trivial narrative developments.

    \item \textbf{Complexity (Comp.):} measures the elaborateness of the narrative in terms of plot structure, character development, and overall sophistication.
\end{itemize}

\paragraph{Annotators.}
The evaluation was conducted by volunteer annotators from our university, representing diverse demographic backgrounds. Each book was independently evaluated by two annotators, both fluent in the language of the evaluated text. Annotators received detailed instructions describing the task and evaluation criteria. Participation was voluntary and uncompensated, and all annotators were informed about the purpose of the study and their role in the evaluation.

\paragraph{Annotation task.}
Given the length of the generated books, five passages were randomly sampled from each book while preserving the narrative flow. Annotators answered five questions per passage and five book-level questions, for a total of $30$ questions per book, using a $5$-point Likert scale. Results are reported in \reftab{tab:human_eval}. The evaluation provides a limited assessment of the linguistic and narrative quality of the generated books, with potential confounding factors including annotator subjectivity and the restricted coverage of models and genres.

\paragraph{Results.}
\reftab{tab:human_eval} reports the average passage-level and overall scores. The generated books achieve strong passage-level results for relevance (4.66), coherence (4.45), fluency (4.15), and diversity (4.18), while engagement is lower (3.73). At the overall level, coherence receives the highest score (4.16), whereas the more subjective dimensions of engagement (3.25) and surprise (3.16) are weaker, indicating more predictable narratives. These results complement the automatic textual analysis (\S\ref{sec:quant-analysis}) and indicate that the generated passages are generally fluent, coherent, diverse, and aligned with the intended narrative constraints. However, the results vary across languages. In particular, Russian books obtain lower ratings across several dimensions but, importantly, maintain high relevance ($4.9$).
Inter-annotator agreement remains low (avg. $0.214$) for a highly subjective task, varying across languages. 
Nevertheless, most paired ratings remain close across languages on avg. $84.4$\%
by at most one point. The limited scale of the evaluation prevents broader language-specific conclusions. Overall, the results indicate adequate linguistic and narrative quality of books in \ourdataset.

\begin{table*}[htbp]
    \centering

    \resizebox{\textwidth}{!}{%
    \begin{tabular}{lcccccc@{\hspace{8pt}}cccccc}
        \toprule[1.5pt]
        \multirow{2}{*}{\textbf{Method}}
        & \multicolumn{6}{c}{\textbf{Low Resource}}
        & \multicolumn{6}{c}{\textbf{High Resource}} \\

        \cmidrule(lr){2-7}
        \cmidrule(lr){8-13}

        & \textbf{\en{}}
        & \textbf{\de{}}
        & \textbf{\ita{}}
        & \textbf{\es{}}
        & \textbf{\rus{}}
        & \textbf{\zh{}}
        & \textbf{\en{}}
        & \textbf{\de{}}
        & \textbf{\ita{}}
        & \textbf{\es{}}
        & \textbf{\rus{}}
        & \textbf{\zh{}} \\
        \midrule

        \rank
        & 0.316 & 0.440 & 0.332 & 0.408 & 0.420 & 0.420
        & 0.340 & 0.488 & 0.240 & 0.460 & 0.340 & 0.332 \\

        \entropy
        & 0.332 & 0.472 & 0.336 & 0.396 & 0.332 & 0.332
        & 0.336 & 0.492 & 0.240 & 0.436 & 0.368 & 0.360 \\

        \gltr
        & 0.352 & 0.408 & 0.336 & 0.424 & 0.436 & 0.320
        & 0.372 & 0.488 & 0.256 & 0.436 & 0.304 & 0.372 \\

        \midrule

        \ngram
        & 0.577 & 0.433 & 0.514 & 0.667 & 0.325 & 0.470
        & 0.757 & 0.937 & 0.946 & 0.964 & 0.658 & 0.874 \\

        \xlmroberta
        & 0.610 & 0.660 & 0.520 & 0.490 & 0.610 & 0.600
        & 0.660 & 0.860 & 0.490 & 0.670 & 0.670 & 0.740 \\

        \detective
        & 0.840 & 0.720 & 0.680 & 0.850 & 0.658 & 0.740
        & 0.890 & 0.850 & 0.630 & 0.690 & 0.950 & 0.930 \\

        \midrule

        \tracerank
        & 0.946 & 0.916 & 0.940 & 0.914 & 0.900 & 0.922
        & 0.945 & 0.918 & 0.937 & 0.927 & 0.934 & 0.930 \\

        \traceentjs
        & 0.900 & 0.933 & 0.957 & 0.939 & 0.911 & 0.953
        & 0.965 & 0.954 & 0.935 & 0.965 & 0.955 & 0.961 \\

        \traceentnor
        & -2.715 & -4.460 & -2.505 & -2.680 & -3.125 & -2.710
        & -2.300 & -2.840 & -3.065 & -2.265 & -2.175 & -2.335 \\

        \bottomrule[1.5pt]
    \end{tabular}%
    }

    \caption{
        Thresholds for each method and language under the low- and
        high-resource settings.
    }
    \label{tab:thresholds}
\end{table*}

\subsection{Cost Analysis}
\refapptab{tab:dataset-API-cost-breakdown} lists generation costs per LLM. In total, generating \ourdataset cost approximately \$$972$.

\begin{table}[H]
    \centering
    \begin{tabular}{lSSS}
        \toprule[1.5pt]
        \multirow{2}{*}{\textbf{LLM}} & \multicolumn{2}{c}{API Cost (\$/$1$M)}  & {\textbf{Total}} \\
        \cmidrule(lr){2-3}
            & \multicolumn{1}{c}{Input}  & \multicolumn{1}{c}{Output} & {(in \$)} \\
        \midrule
        \geminiPro   & 1.25 & 10.00 & 760.0 \\
        \geminiFlash & 0.30 & 2.50  & 61.6  \\
        \deepseek    & 0.20 & 0.31  & 58.1  \\
        \qwen        & 0.09 & 0.55  & 45.7  \\
        \gptOSS      & 0.03 & 0.17  & 46.4  \\
         \midrule
         \multicolumn{3}{c}{Total} & 971.8 \\
         \bottomrule[1.5pt]
    \end{tabular}
    \caption{\ourdataset generation cost breakdown. `Input' and `Output' denote API cost per $1$M tokens. `Total' is the cost of generating all books for a given LLM. \texttt{OpenRouter} API costs are as of April 2026. }
    \label{tab:dataset-API-cost-breakdown}
\end{table}

\section{Authorship Attribution Methods (cont.)}
\label{app:authorship-attribution-methods}

\paragraph{\rank.} For each text, we compute the average rank of the observed tokens.

\paragraph{\entropy.} Similarly, we compute the average token-level entropy.

\paragraph{\gltr.} \gltr~\citep{gehrmann-etal-2019-gltr} represents a text through the proportion of tokens falling into four predefined rank buckets.

For all three metric-based detectors, we follow \citet{OTB,ghostwritebench} and adapt the extracted statistics to multi-class attribution by training a logistic regression classifier to predict the generating LLM. Following \citet{multitude}, we use \mgpt~\citep{mgpt} as the multilingual reference language model.

\paragraph{\ngram.} \ngram~\citep{koppel2004authorship} uses three complementary feature representations. The \textit{char} analyser preserves letters, punctuation, and spaces. The \textit{dist\_char} analyser replaces letters with a common placeholder, retaining structural patterns involving punctuation, numbers, symbols, and spacing. The \textit{word} analyser removes punctuation and operates on word-level features. To preserve language-specific information, accented and language-specific characters are retained during preprocessing. For Chinese, word-level features are extracted after segmenting the text with \textit{Jieba}.\footnote{\url{https://pypi.org/project/jieba/}}

\paragraph{\bert.} A multi-class classification layer is added on top of the pre-trained encoder, and the complete architecture is fine-tuned to predict the generating LLM. In our implementation, the underlying encoder is \xlmroberta~\citep{conneau-etal-2020-unsupervised}.

\paragraph{\detective.} \detective~\citep{guo2024detective} relies on sentence embeddings learned through a multi-level contrastive objective. For AA, these sentence embeddings are used to train a multi-class classifier over the generating LLMs. The original sentence-embedding model is replaced with \path{ZurichNLP/unsup-simcse-xlm-roberta-base},\footnote{\url{https://huggingface.co/ZurichNLP/unsup-simcse-xlm-roberta-base}} its multilingual counterpart. 
We also considered the more recent \textsc{M-RangeDetector} \citep{rangedetector} and simpler contrastive-learning approaches \citep{malyalam-contrastive-AA}. However, their implementations are not publicly available, and the methodological details provided were insufficient for reliable reproduction.

\paragraph{\trace.} For each text, \trace~\citep{ghostwritebench} models transitions between consecutive token-rank or entropy values, producing a two-dimensional fingerprint. A reference fingerprint is constructed for each generator from its training texts, and a test text is attributed to the generator whose reference fingerprint is most similar to the fingerprint extracted from the test text. \tracerank uses rank-based transition fingerprints, whereas \traceentjs and \traceentnor use entropy-based fingerprints and compare them through Jensen-Shannon divergence and norm-based distance, respectively. We use \gemma\footnote{\url{https://huggingface.co/google/gemma-3-1b-it}} as the evaluator language model.

\section{Experimental Details}
\label{app:impl_det}
We use OpenRouter\footnote{\url{https://openrouter.ai/}} for all LLM API calls used to generate the benchmark. We adopt the same decoding parameters for all generators, setting \texttt{temperature} to $1.0$, \texttt{top-p} to $1.0$, \texttt{top-k} to $0$, and \texttt{maximum\_output\_length} to $8192$ tokens. We provide a model card for each LLM used in this work in \reftab{tab:llm-model-card}. All experiments were conducted using a single \texttt{A100 GPU} with \texttt{CUDA 12.4} and \texttt{PyTorch 2.10.0}.

Since \xlmroberta cannot process complete long-form documents directly, we split each text into chunks of $512$ tokens and average the chunk-level prediction scores to obtain the final document-level prediction, following \citet{valla, ghostwritebench}. For \detective, since it is clustering-based, we compute the proportion of top-$K$ results assigned to each generator across the text chunks.

All hyperparameter tuning was performed on the development set. The number of fine-tuning epochs depends on the experimental configuration. For \xlmroberta, we fine-tune the model for up to $10$ epochs and retain the checkpoint achieving the highest macro-F1 on the development set. For \detective, we train for up to $50$ epochs and similarly select the checkpoint with the best development-set macro-F1. For all \trace variants, we follow the hyperparameter configuration of \citet{ghostwritebench}. Specifically, for \tracerank, we set $\alpha=1.5$ and use $100$K samples to approximate the power-law distribution. We use \gemma as the evaluator language model with a context length of $1024$ tokens. The entropy-based variants use a grid size of $50$, whereas \tracerank uses $50$ clusters.

\section{Additional Results}

\begin{figure*}[!htp]
    \centering
    \includegraphics[
        width=\textwidth,
        trim={0 0 0 0},
        clip
    ]{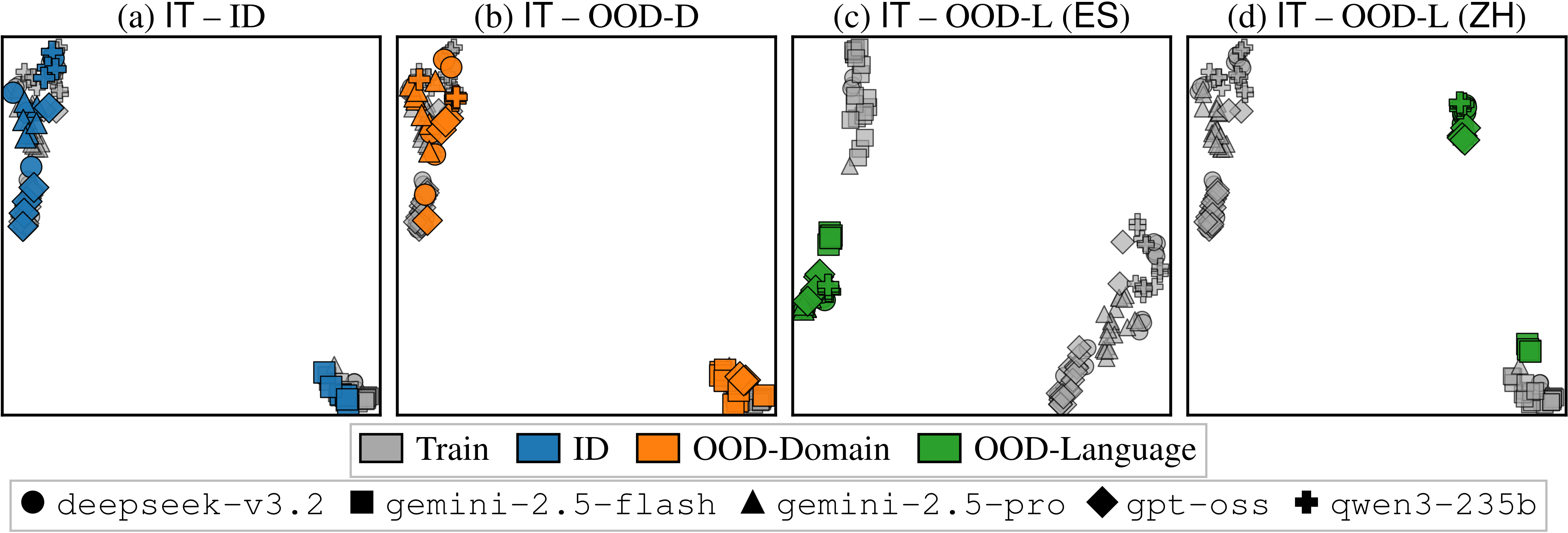}
    \caption{
    UMAP projections of \trace entropy fingerprints using Italian as the
    reference language.
    The panels compare Italian training fingerprints with (a) Italian ID-test,
    (b) Italian \oodOne, and (c) Spanish and (d) Chinese ID fingerprints evaluated
    under \oodThree.
    Grey points denote the Italian training references, while blue, orange,
    and green points denote ID, \oodOne, and cross-language fingerprints,
    respectively.
    Marker shapes identify the LLM generator.
    Italian ID-test and \oodOne fingerprints remain close to the training
    references, whereas both Spanish and Chinese fingerprints occupy clearly
    displaced regions, indicating that \trace fingerprints are strongly
    language-dependent and transfer poorly under \oodThree.
    }
    \label{fig:trace_reference_comparison}
\end{figure*}

This section provides additional analyses and complete results for all evaluation settings. \reffig{fig:trace_reference_comparison} extends the qualitative analysis of \trace by comparing entropy fingerprints from Italian training texts with Italian ID, Italian \oodOne, and Spanish and Chinese \oodThree texts. \reftab{tab:results_low} and \reftab{tab:in-lang-results-high} report the threshold-based in-language results for ID, \oodOne, and \oodTwo in the low- and high-resource settings, while \reftab{tab:low_resource_deltas} and \reftab{tab:high_resource_deltas} report the corresponding changes in macro-F1 relative to ID performance. \reftab{tab:results_low_without_threshold} and \reftab{tab:results_high_without_threshold} provide the ID and \oodOne results without threshold-based rejection. Finally, \reftab{tab:ood_language_threshold_all} and \reftab{tab:ood_language_no_threshold_all} report \oodThree results across all source--target language pairs, with and without thresholding, under both resource settings.

\begin{sidewaystable*}[p]
\centering
\scriptsize

\begin{tabular}{l|ccc|ccc|ccc|ccc|ccc|ccc}
\toprule[1.5pt]
& \multicolumn{3}{c}{\textbf{\en{}}}
& \multicolumn{3}{c}{\textbf{\de{}}}
& \multicolumn{3}{c}{\textbf{\ita{}}}
& \multicolumn{3}{c}{\textbf{\es{}}}
& \multicolumn{3}{c}{\textbf{\rus{}}}
& \multicolumn{3}{c}{\textbf{\zh{}}} \\
\cmidrule(lr){2-4}
\cmidrule(lr){5-7}
\cmidrule(lr){8-10}
\cmidrule(lr){11-13}
\cmidrule(lr){14-16}
\cmidrule(lr){17-19}

\textbf{Method}
& \textbf{ID} & \textbf{OOD-D} & \textbf{OOD-A}
& \textbf{ID} & \textbf{OOD-D} & \textbf{OOD-A}
& \textbf{ID} & \textbf{OOD-D} & \textbf{OOD-A}
& \textbf{ID} & \textbf{OOD-D} & \textbf{OOD-A}
& \textbf{ID} & \textbf{OOD-D} & \textbf{OOD-A}
& \textbf{ID} & \textbf{OOD-D} & \textbf{OOD-A} \\
\midrule

\rank
& 0.245 & 0.047 & 0
& 0.114 & 0.191 & 0.080
& 0 & 0 & 0.105
& 0.080 & 0.066 & 0
& 0.094 & 0.080 & 0.150
& 0.066 & 0.200 & 0.388 \\

\entropy
& 0.250 & 0.227 & 0.028
& 0 & 0 & 0.300
& 0.111 & 0.336 & 0
& 0.120 & 0.107 & 0.176
& 0.286 & 0.116 & 0
& 0.290 & 0.289 & 0.072 \\

\gltr
& 0.066 & 0.050 & 0
& 0.090 & 0.070 & 0
& 0.080 & 0 & 0
& 0.100 & 0.100 & 0.200
& 0.111 & 0.076 & 0
& 0.220 & 0.274 & 0 \\

\midrule

\ngram
& 0.447 & 0.335 & 0.500
& 0.639 & 0.634 & 0.309
& 0.493 & 0.277 & 0.060
& 0.581 & 0.600 & \textbf{0.974}
& 0.605 & \underline{0.517} & 0.040
& \textbf{0.960} & 0.792 & \textbf{0.843} \\

\xlmroberta
& \textbf{0.700} & \textbf{0.531} & 0.485
& 0.620 & 0.593 & \textbf{0.548}
& \textbf{0.842} & 0.426 & 0.317
& \textbf{0.886} & \underline{0.654} & 0.280
& \underline{0.661} & \textbf{0.630} & 0.429
& 0.882 & \underline{0.872} & 0.696 \\

\detective
& 0.494 & 0.398 & 0.527
& 0.511 & 0.525 & \underline{0.323}
& 0.512 & 0.341 & 0.172
& 0.665 & 0.515 & 0.505
& 0.506 & 0.381 & 0.327
& 0.891 & 0.766 & 0.487 \\

\midrule

\tracerank
& 0.486 & 0.271 & \underline{0.614}
& 0.479 & 0.396 & 0.080
& \underline{0.709} & \textbf{0.500} & 0.484
& 0.391 & 0.460 & 0.206
& \textbf{0.718} & 0.516 & 0.033
& \underline{0.898} & 0.754 & 0.328 \\

\traceentjs
& 0.587 & 0.472 & 0.080
& \underline{0.868} & \underline{0.702} & 0.246
& 0.588 & 0.374 & \underline{0.660}
& \underline{0.801} & \textbf{0.707} & 0.501
& 0.566 & 0.302 & \underline{0.783}
& 0.791 & 0.856 & \underline{0.744} \\

\traceentnor
& \underline{0.641} & \underline{0.489} & \textbf{0.719}
& \textbf{0.916} & \textbf{0.790} & 0.030
& 0.540 & \underline{0.435} & \textbf{0.740}
& 0.774 & 0.542 & \underline{0.774}
& 0.413 & 0.253 & \textbf{0.888}
& 0.829 & \textbf{0.892} & 0.650 \\

\bottomrule[1.5pt]
\end{tabular}%

\caption[Low-resource threshold-based results]
{Threshold-based low-resource results for ID, \oodOne, and \oodTwo evaluation in each language.}
\label{tab:results_low}
\end{sidewaystable*}

\begin{sidewaystable*}[p]
\centering
\scriptsize

\begin{tabular}{l|ccc|ccc|ccc|ccc|ccc|ccc}
\toprule[1.5pt]
& \multicolumn{3}{c}{\textbf{\en{}}}
& \multicolumn{3}{c}{\textbf{\de{}}}
& \multicolumn{3}{c}{\textbf{\ita{}}}
& \multicolumn{3}{c}{\textbf{\es{}}}
& \multicolumn{3}{c}{\textbf{\rus{}}}
& \multicolumn{3}{c}{\textbf{\zh{}}} \\
\cmidrule(lr){2-4}
\cmidrule(lr){5-7}
\cmidrule(lr){8-10}
\cmidrule(lr){11-13}
\cmidrule(lr){14-16}
\cmidrule(lr){17-19}

\textbf{Method}
& \textbf{ID} & \textbf{OOD-D} & \textbf{OOD-A}
& \textbf{ID} & \textbf{OOD-D} & \textbf{OOD-A}
& \textbf{ID} & \textbf{OOD-D} & \textbf{OOD-A}
& \textbf{ID} & \textbf{OOD-D} & \textbf{OOD-A}
& \textbf{ID} & \textbf{OOD-D} & \textbf{OOD-A}
& \textbf{ID} & \textbf{OOD-D} & \textbf{OOD-A} \\
\midrule

\rank
& 0 & 0.142 & 0.200
& 0.057 & 0.125 & 0.317
& 0.095 & 0.046 & 0
& 0.080 & 0.066 & 0
& 0.166 & 0.093 & 0.114
& 0.114 & 0.118 & 0 \\

\entropy
& 0.070 & 0.120 & 0.120
& 0.023 & 0 & 0.200
& 0.180 & 0.129 & 0
& 0.120 & 0.107 & 0.120
& 0.192 & 0.100 & 0.036
& 0.277 & 0.303 & 0.156 \\

\gltr
& 0.040 & 0.168 & 0.200
& 0.106 & 0.149 & 0.257
& 0.197 & 0.308 & 0
& 0.277 & 0.150 & 0.164
& 0.200 & 0.267 & 0
& 0.125 & 0.278 & 0.364 \\

\midrule

\ngram
& \textbf{1.000} & \textbf{1.000} & \textbf{0.980}
& 0.803 & 0.659 & \textbf{1.000}
& 0.896 & \underline{0.725} & \textbf{1.000}
& 0.869 & 0.633 & \textbf{0.992}
& \textbf{1.000} & \textbf{0.966} & \underline{0.733}
& \underline{0.971} & \underline{0.959} & \textbf{0.946} \\

\xlmroberta
& \textbf{1.000} & \underline{0.924} & 0.334
& 0.949 & \underline{0.775} & 0.660
& \textbf{0.956} & 0.676 & 0.203
& \textbf{1.000} & \textbf{0.933} & 0.441
& \underline{0.977} & \underline{0.960} & 0.474
& \textbf{1.000} & \textbf{1.000} & 0.653 \\

\detective
& 0.931 & 0.786 & 0.681
& \underline{0.955} & 0.658 & \underline{0.707}
& \underline{0.924} & 0.677 & \underline{0.477}
& \underline{0.953} & \underline{0.864} & 0.518
& 0.769 & 0.717 & \textbf{0.761}
& 0.844 & 0.845 & 0.866 \\

\midrule

\tracerank
& 0.682 & 0.660 & 0.303
& 0.694 & 0.490 & 0.100
& 0.831 & 0.684 & 0.189
& 0.361 & 0.466 & 0.357
& 0.959 & 0.501 & 0.197
& 0.801 & 0.838 & 0.285 \\

\traceentjs
& 0.878 & 0.541 & \underline{0.732}
& \textbf{0.959} & \textbf{0.905} & 0.537
& 0.810 & 0.662 & 0.050
& 0.736 & 0.610 & \underline{0.963}
& \textbf{1.000} & 0.741 & 0.322
& 0.933 & 0.814 & 0.683 \\

\traceentnor
& \underline{0.966} & 0.755 & 0.709
& \textbf{0.959} & \textbf{0.905} & 0.405
& 0.893 & \textbf{0.793} & 0.164
& 0.869 & 0.624 & 0.922
& 0.931 & 0.737 & 0.504
& 0.880 & 0.945 & 0.592 \\

\bottomrule[1.5pt]
\end{tabular}%

\caption{Threshold-based high-resource results for ID, \oodOne, and \oodTwo evaluation in each language.}
\label{tab:in-lang-results-high}
\end{sidewaystable*}

\begin{sidewaystable*}[p]
\centering

\begin{tabular}{l|cc|cc|cc|cc|cc|cc}
\toprule[1.5pt]
& \multicolumn{2}{c}{\textbf{\en{}}}
& \multicolumn{2}{c}{\textbf{\de{}}}
& \multicolumn{2}{c}{\textbf{\ita{}}}
& \multicolumn{2}{c}{\textbf{\es{}}}
& \multicolumn{2}{c}{\textbf{\rus{}}}
& \multicolumn{2}{c}{\textbf{\zh{}}} \\
\cmidrule(lr){2-3}
\cmidrule(lr){4-5}
\cmidrule(lr){6-7}
\cmidrule(lr){8-9}
\cmidrule(lr){10-11}
\cmidrule(lr){12-13}

\textbf{Method}
& \textbf{ID} & \textbf{OOD-D}
& \textbf{ID} & \textbf{OOD-D}
& \textbf{ID} & \textbf{OOD-D}
& \textbf{ID} & \textbf{OOD-D}
& \textbf{ID} & \textbf{OOD-D}
& \textbf{ID} & \textbf{OOD-D} \\
\midrule

\rank
& 0.261 & 0.045
& 0.209 & 0.191
& 0.140 & 0.007
& 0.075 & 0.068
& 0.074 & 0.066
& 0.071 & 0.070 \\

\entropy
& 0.248 & 0.207
& 0.068 & 0.063
& 0.085 & 0.294
& 0.075 & 0.069
& 0.286 & 0.116
& 0.231 & 0.247 \\

\gltr
& 0.050 & 0.052
& 0.071 & 0.064
& 0.077 & 0.078
& 0.075 & 0.068
& 0.074 & 0.066
& 0.227 & 0.236 \\

\midrule

\ngram
& 0.476 & 0.483
& 0.625 & 0.616
& 0.490 & 0.374
& 0.699 & 0.627
& 0.591 & \underline{0.582}
& \textbf{1.000} & \underline{0.906} \\

\xlmroberta
& \textbf{0.780} & \underline{0.531}
& 0.696 & 0.595
& \textbf{0.842} & \underline{0.456}
& \textbf{0.893} & 0.666
& \textbf{0.814} & 0.566
& 0.876 & 0.874 \\

\detective
& 0.476 & 0.475
& 0.500 & 0.600
& 0.482 & 0.305
& 0.617 & 0.650
& 0.488 & 0.470
& 0.915 & 0.826 \\

\midrule

\tracerank
& 0.440 & 0.294
& 0.479 & 0.376
& \underline{0.694} & \textbf{0.579}
& 0.395 & 0.420
& 0.718 & 0.509
& 0.926 & 0.787 \\

\traceentjs
& 0.587 & 0.469
& \underline{0.868} & \underline{0.702}
& 0.557 & 0.408
& 0.786 & \textbf{0.724}
& 0.708 & 0.328
& 0.861 & 0.876 \\

\traceentnor
& \underline{0.621} & \textbf{0.562}
& \textbf{0.916} & \textbf{0.790}
& 0.579 & 0.446
& \underline{0.875} & \underline{0.699}
& \underline{0.733} & \textbf{0.594}
& \underline{0.949} & \textbf{0.920} \\

\bottomrule[1.5pt]
\end{tabular}%

\caption{Low-resource results without threshold-based rejection for ID and \oodOne evaluation in each language.}
\label{tab:results_low_without_threshold}
\end{sidewaystable*}

\begin{sidewaystable*}[p]
\centering

\begin{tabular}{l|cc|cc|cc|cc|cc|cc}
\toprule[1.5pt]
& \multicolumn{2}{c}{\textbf{\en{}}}
& \multicolumn{2}{c}{\textbf{\de{}}}
& \multicolumn{2}{c}{\textbf{\ita{}}}
& \multicolumn{2}{c}{\textbf{\es{}}}
& \multicolumn{2}{c}{\textbf{\rus{}}}
& \multicolumn{2}{c}{\textbf{\zh{}}} \\
\cmidrule(lr){2-3}
\cmidrule(lr){4-5}
\cmidrule(lr){6-7}
\cmidrule(lr){8-9}
\cmidrule(lr){10-11}
\cmidrule(lr){12-13}

\textbf{Method}
& \textbf{ID} & \textbf{OOD-D}
& \textbf{ID} & \textbf{OOD-D}
& \textbf{ID} & \textbf{OOD-D}
& \textbf{ID} & \textbf{OOD-D}
& \textbf{ID} & \textbf{OOD-D}
& \textbf{ID} & \textbf{OOD-D} \\
\midrule

\rank
& 0.070 & 0.078
& 0.068 & 0.060
& 0.070 & 0.055
& 0.075 & 0.068
& 0.074 & 0.082
& 0.082 & 0.082 \\

\entropy
& 0.129 & 0.080
& 0.068 & 0.063
& 0.180 & 0.150
& 0.075 & 0.167
& 0.228 & 0.169
& 0.300 & 0.300 \\

\gltr
& 0.070 & 0.078
& 0.068 & 0.063
& 0.347 & 0.352
& 0.220 & 0.223
& 0.348 & 0.313
& 0.082 & 0.236 \\

\midrule

\ngram
& \textbf{1.000} & \textbf{1.000}
& \textbf{1.000} & \textbf{1.000}
& \textbf{1.000} & \textbf{0.974}
& \textbf{1.000} & \textbf{0.972}
& \textbf{1.000} & \textbf{1.000}
& \textbf{1.000} & \textbf{1.000} \\

\xlmroberta
& \textbf{1.000} & \textbf{1.000}
& 0.953 & 0.819
& \underline{0.956} & 0.676
& \textbf{1.000} & \underline{0.946}
& \textbf{1.000} & \underline{0.968}
& \textbf{1.000} & \textbf{1.000} \\

\detective
& \textbf{1.000} & \textbf{1.000}
& \textbf{1.000} & \underline{0.850}
& 0.924 & 0.673
& \underline{0.966} & \underline{0.946}
& \textbf{1.000} & \underline{0.968}
& \textbf{1.000} & \underline{0.971} \\

\midrule

\tracerank
& \underline{0.719} & 0.771
& 0.694 & 0.536
& 0.831 & 0.734
& 0.415 & 0.550
& \underline{0.959} & 0.494
& 0.847 & 0.888 \\

\traceentjs
& \textbf{1.000} & 0.710
& \underline{0.959} & \underline{0.905}
& 0.810 & 0.662
& 0.919 & 0.800
& \textbf{1.000} & 0.741
& \underline{0.949} & 0.935 \\

\traceentnor
& \textbf{1.000} & \underline{0.850}
& \underline{0.959} & \underline{0.905}
& 0.893 & \underline{0.793}
& 0.919 & 0.800
& \underline{0.959} & 0.690
& \textbf{1.000} & \textbf{1.000} \\

\bottomrule[1.5pt]
\end{tabular}%

\caption[High-resource results without threshold]
{High-resource results without threshold-based rejection for ID and \oodOne evaluation in each language.}
\label{tab:results_high_without_threshold}
\end{sidewaystable*}

\begin{table*}[t]
\centering
\resizebox{\textwidth}{!}{%
\begin{tabular}{
l
*{12}{S[table-format=+1.3]}
@{\hspace{0.45cm}}
*{2}{S[table-format=+1.3]}
}
\toprule
\multirow{2}{*}{\textbf{Detector}}
& \multicolumn{2}{c}{\textbf{EN}}
& \multicolumn{2}{c}{\textbf{DE}}
& \multicolumn{2}{c}{\textbf{IT}}
& \multicolumn{2}{c}{\textbf{ES}}
& \multicolumn{2}{c}{\textbf{RU}}
& \multicolumn{2}{c}{\textbf{ZH}}
& \multicolumn{2}{c}{\textbf{Avg.}} \\

\cmidrule(lr){2-3}
\cmidrule(lr){4-5}
\cmidrule(lr){6-7}
\cmidrule(lr){8-9}
\cmidrule(lr){10-11}
\cmidrule(lr){12-13}
\cmidrule(lr){14-15}

& {$\Delta_{\mathrm{D}}$} & {$\Delta_{\mathrm{A}}$}
& {$\Delta_{\mathrm{D}}$} & {$\Delta_{\mathrm{A}}$}
& {$\Delta_{\mathrm{D}}$} & {$\Delta_{\mathrm{A}}$}
& {$\Delta_{\mathrm{D}}$} & {$\Delta_{\mathrm{A}}$}
& {$\Delta_{\mathrm{D}}$} & {$\Delta_{\mathrm{A}}$}
& {$\Delta_{\mathrm{D}}$} & {$\Delta_{\mathrm{A}}$}
& {$\Delta_{\mathrm{D}}$} & {$\Delta_{\mathrm{A}}$} \\
\midrule

\rank
& -0.198 & -0.245
& +0.077 & -0.034
&  0.000 & +0.105
& -0.014 & -0.080
& -0.014 & +0.056
& +0.134 & +0.322
& -0.003 & +0.021 \\

\entropy
& -0.023 & -0.222
&  0.000 & +0.300
& +0.225 & -0.111
& -0.013 & +0.056
& -0.170 & -0.286
& -0.001 & -0.218
& +0.003 & -0.080 \\

\gltr
& -0.016 & -0.066
& -0.020 & -0.090
& -0.080 & -0.080
&  0.000 & +0.100
& -0.035 & -0.111
& +0.054 & -0.220
& -0.016 & -0.078 \\

\midrule

\ngram
& -0.112 & +0.053
& -0.005 & -0.330
& -0.216 & -0.433
& +0.019 & +0.393
& -0.088 & -0.565
& -0.168 & -0.117
& -0.095 & -0.166 \\

\xlmroberta
& -0.169 & -0.215
& -0.027 & -0.072
& -0.416 & -0.525
& -0.232 & -0.606
& -0.031 & -0.232
& -0.010 & -0.186
& -0.148 & -0.306 \\

\detective
& -0.096 & +0.033
& +0.014 & -0.188
& -0.171 & -0.340
& -0.150 & -0.160
& -0.125 & -0.179
& -0.125 & -0.404
& -0.109 & -0.206 \\

\midrule

\tracerank
& -0.215 & +0.128
& -0.083 & -0.399
& -0.209 & -0.225
& +0.069 & -0.185
& -0.202 & -0.685
& -0.144 & -0.570
& -0.131 & -0.323 \\

\traceentjs
& -0.115 & -0.507
& -0.166 & -0.622
& -0.214 & +0.072
& -0.094 & -0.300
& -0.264 & +0.217
& +0.065 & -0.047
& -0.131 & -0.198 \\

\traceentnor
& -0.152 & +0.078
& -0.126 & -0.886
& -0.105 & +0.200
& -0.232 &  0.000
& -0.160 & +0.475
& +0.063 & -0.179
& -0.119 & -0.052 \\

\bottomrule
\end{tabular}%
}

\caption{
Changes in macro-F1 under OOD evaluation relative to ID performance in the
low-resource setting.
For each detector and language, the table reports
$\Delta_{\mathrm{D}} =
F_1^{\mathrm{OOD\text{-}D}} - F_1^{\mathrm{ID}}$
and
$\Delta_{\mathrm{A}} =
F_1^{\mathrm{OOD\text{-}A}} - F_1^{\mathrm{ID}}$.
Negative values indicate degradation, while positive values indicate improvement.
The Avg.\ columns report averages across languages.
}
\label{tab:low_resource_deltas}
\end{table*}

\begin{table*}[t]
\centering
\resizebox{\textwidth}{!}{%
\begin{tabular}{
l
*{12}{S[table-format=+1.3]}
@{\hspace{0.45cm}}
*{2}{S[table-format=+1.3]}
}
\toprule
\multirow{2}{*}{\textbf{Detector}}
& \multicolumn{2}{c}{\textbf{EN}}
& \multicolumn{2}{c}{\textbf{DE}}
& \multicolumn{2}{c}{\textbf{IT}}
& \multicolumn{2}{c}{\textbf{ES}}
& \multicolumn{2}{c}{\textbf{RU}}
& \multicolumn{2}{c}{\textbf{ZH}}
& \multicolumn{2}{c}{\textbf{Avg.}} \\

\cmidrule(lr){2-3}
\cmidrule(lr){4-5}
\cmidrule(lr){6-7}
\cmidrule(lr){8-9}
\cmidrule(lr){10-11}
\cmidrule(lr){12-13}
\cmidrule(lr){14-15}

& {$\Delta_{\mathrm{D}}$} & {$\Delta_{\mathrm{A}}$}
& {$\Delta_{\mathrm{D}}$} & {$\Delta_{\mathrm{A}}$}
& {$\Delta_{\mathrm{D}}$} & {$\Delta_{\mathrm{A}}$}
& {$\Delta_{\mathrm{D}}$} & {$\Delta_{\mathrm{A}}$}
& {$\Delta_{\mathrm{D}}$} & {$\Delta_{\mathrm{A}}$}
& {$\Delta_{\mathrm{D}}$} & {$\Delta_{\mathrm{A}}$}
& {$\Delta_{\mathrm{D}}$} & {$\Delta_{\mathrm{A}}$} \\
\midrule

\rank
& +0.142 & +0.200
& +0.068 & +0.260
& -0.049 & -0.095
& -0.014 & +0.040
& -0.073 & -0.052
& +0.004 & +0.042
& +0.013 & +0.066 \\

\entropy
& +0.050 & +0.130
& -0.023 & -0.023
& -0.051 & -0.180
& -0.013 & +0.044
& -0.079 & -0.139
& +0.026 & +0.087
& -0.015 & -0.014 \\

\gltr
& +0.128 & +0.160
& +0.043 & +0.151
& +0.111 & -0.197
& -0.127 & -0.113
& +0.067 & -0.200
& +0.153 & -0.079
& +0.063 & -0.046 \\

\midrule

\ngram
&  0.000 & -0.020
& -0.144 & +0.197
& -0.171 & +0.104
& -0.236 & +0.123
& -0.034 & -0.268
& -0.012 & -0.025
& -0.100 & +0.019 \\

\xlmroberta
& -0.076 & -0.666
& -0.174 & -0.289
& -0.280 & -0.753
& -0.067 & -0.559
& -0.071 & -0.318
&  0.000 & -0.317
& -0.111 & -0.484 \\

\detective
& -0.145 & -0.250
& -0.297 & -0.248
& -0.247 & -0.447
& -0.089 & -0.435
& -0.052 & -0.008
& +0.001 & +0.022
& -0.138 & -0.228 \\

\midrule

\tracerank
& -0.230 & -0.640
& -0.182 & -0.692
& -0.147 & -0.642
& +0.105 & -0.004
& -0.200 & -0.679
& +0.037 & -0.516
& -0.103 & -0.529 \\

\traceentjs
& -0.337 & -0.146
& -0.054 & -0.422
& -0.148 & -0.760
& -0.126 & +0.227
& -0.259 & -0.037
& -0.119 & -0.250
& -0.174 & -0.231 \\

\traceentnor
& -0.211 & -0.257
& -0.054 & -0.554
& -0.100 & -0.729
& -0.245 & +0.053
& -0.194 & -0.427
& +0.065 & -0.288
& -0.123 & -0.367 \\

\bottomrule
\end{tabular}%
}

\caption{
Changes in macro-F1 under OOD evaluation relative to ID performance in the
high-resource setting.
For each detector and language, the table reports
$\Delta_{\mathrm{D}} =
F_1^{\mathrm{OOD\text{-}D}} - F_1^{\mathrm{ID}}$
and
$\Delta_{\mathrm{A}} =
F_1^{\mathrm{OOD\text{-}A}} - F_1^{\mathrm{ID}}$.
Negative values indicate degradation, while positive values indicate improvement.
The Avg.\ columns report averages across languages.
}
\label{tab:high_resource_deltas}
\end{table*}

\begin{table*}[p]
\centering
\small
\resizebox{\textwidth}{!}{%
\begin{tabular}{llcccccc@{\hspace{0.45cm}}cccccc}
\toprule[1.5pt]
\multirow{2}{*}{\textbf{Train}} 
& \multirow{2}{*}{\textbf{Method}}
& \multicolumn{6}{c}{\textbf{Low Resource}}
& \multicolumn{6}{c}{\textbf{High Resource}} \\
\cmidrule(lr){3-8}
\cmidrule(lr){9-14}
& 
& \textbf{\en{}} & \textbf{\de{}} & \textbf{\ita{}} & \textbf{\es{}} & \textbf{\rus{}} & \textbf{\zh{}}
& \textbf{\en{}} & \textbf{\de{}} & \textbf{\ita{}} & \textbf{\es{}} & \textbf{\rus{}} & \textbf{\zh{}} \\
\midrule

\multirow{9}{*}{\en{}}
& \rank
& -- & 0.068 & 0.055 & 0.053 & 0.061 & 0.071
& -- & 0 & 0 & 0 & 0 & 0 \\
& \entropy
& -- & 0.068 & 0.055 & 0.053 & 0.061 & 0.071
& -- & 0.030 & 0.055 & 0.053 & 0.061 & 0 \\
& \gltr
& -- & 0 & 0 & 0.033 & 0.05 & 0.044
& -- & 0.103 & 0 & 0 & 0 & 0 \\
\cmidrule(lr){2-14}
& \ngram
& -- & 0 & 0 & 0 & 0 & 0
& -- & 0 & 0 & 0 & 0 & 0 \\
& \xlmroberta
& -- & \textbf{0.683} & \textbf{0.611} & \textbf{0.642} & \textbf{0.666} & \textbf{0.353}
& -- & \textbf{0.916} & \textbf{0.935} & \textbf{0.914} & \textbf{0.950} & \underline{0.333} \\
& \detective
& -- & \underline{0.542} & \underline{0.527} & \underline{0.450} & \underline{0.477} & \underline{0.094}
& -- & \underline{0.766} & \underline{0.753} & \underline{0.793} & \underline{0.533} & \textbf{0.424} \\
\cmidrule(lr){2-14}
& \tracerank
& -- & 0.112 & 0.404 & 0.295 & 0.203 & 0
& -- & 0 & 0 & 0 & 0 & 0 \\
& \traceentjs
& -- & 0 & 0 & 0 & 0 & 0
& -- & 0 & 0 & 0 & 0 & 0 \\
& \traceentnor
& -- & 0 & 0 & 0 & 0 & 0
& -- & 0 & 0 & 0 & 0 & 0 \\

\midrule

\multirow{9}{*}{\de{}}
& \rank
& 0.070 & -- & 0.068 & 0.100 & 0.061 & 0.080
& 0.070 & -- & 0 & 0 & 0 & 0 \\
& \entropy
& 0 & -- & 0.025 & 0 & 0.074 & 0.033
& 0 & -- & 0.055 & 0.077 & 0.074 & 0.095 \\
& \gltr
& 0.070 & -- & 0.213 & 0.175 & 0.236 & 0.092
& 0.072 & -- & 0 & 0.023 & 0.057 & 0.133 \\
\cmidrule(lr){2-14}
& \ngram
& 0.200 & -- & 0.100 & 0.100 & 0.114 & 0.133
& 0 & -- & 0 & 0 & 0 & 0 \\
& \xlmroberta
& \underline{0.420} & -- & \underline{0.555} & \underline{0.463} & \underline{0.466} & \textbf{0.271}
& \underline{0.673} & -- & \textbf{0.836} & \textbf{0.848} & \textbf{0.800} & \textbf{0.723} \\
& \detective
& \textbf{0.550} & -- & \textbf{0.650} & \textbf{0.602} & \textbf{0.600} & \underline{0.100}
& \textbf{0.766} & -- & \underline{0.772} & \underline{0.617} & 0.394 & \underline{0.559} \\
\cmidrule(lr){2-14}
& \tracerank
& 0.057 & -- & 0.038 & 0.376 & 0.273 & 0
& 0.150 & -- & 0.198 & 0.283 & \underline{0.675} & 0 \\
& \traceentjs
& 0 & -- & 0 & 0 & 0 & 0
& 0 & -- & 0 & 0 & 0 & 0 \\
& \traceentnor
& 0 & -- & 0 & 0.160 & 0.08 & 0
& 0 & -- & 0 & 0 & 0 & 0 \\

\midrule

\multirow{9}{*}{\ita{}}
& \rank
& 0.050 & 0.068 & -- & 0.083 & 0 & 0.025
& 0.050 & 0.071 & -- & 0.179 & 0.064 & 0.094 \\
& \entropy
& 0.071 & 0.090 & -- & 0 & 0.106 & 0.317
& 0.070 & 0.095 & -- & 0.078 & 0.074 & 0.290 \\
& \gltr
& 0.050 & 0.090 & -- & 0.125 & 0.061 & 0.069
& 0.248 & 0.205 & -- & 0.302 & 0.061 & 0.260 \\
\cmidrule(lr){2-14}
& \ngram
& 0.080 & 0.086 & -- & 0.114 & 0.133 & 0.076
& 0 & 0 & -- & 0 & 0 & 0 \\
& \xlmroberta
& \textbf{0.585} & \textbf{0.589} & -- & \textbf{0.750} & \textbf{0.669} & \textbf{0.765}
& \textbf{0.938} & \textbf{0.916} & -- & \textbf{0.981} & \textbf{0.824} & \underline{0.811} \\
& \detective
& \underline{0.371} & \underline{0.468} & -- & \underline{0.471} & \underline{0.206} & \underline{0.304}
& \underline{0.895} & \underline{0.749} & -- & \underline{0.959} & \underline{0.788} & \textbf{0.813} \\
\cmidrule(lr){2-14}
& \tracerank
& 0 & 0 & -- & 0 & 0.08 & 0
& 0 & 0 & -- & 0.189 & 0.05 & 0 \\
& \traceentjs
& 0 & 0 & -- & 0 & 0 & 0
& 0 & 0 & -- & 0 & 0 & 0 \\
& \traceentnor
& 0 & 0 & -- & 0 & 0 & 0
& 0 & 0 & -- & 0 & 0 & 0 \\

\midrule

\multirow{9}{*}{\es{}}
& \rank
& 0.070 & 0.071 & 0.088 & -- & 0.066 & 0.125
& 0.070 & 0.068 & 0.055 & -- & 0.074 & 0.082 \\
& \entropy
& 0.071 & 0.083 & 0.114 & -- & 0 & 0.063
& 0.071 & 0.083 & 0.114 & -- & 0 & 0.063 \\
& \gltr
& 0.070 & 0.088 & 0 & -- & 0.066 & 0.109
& 0.070 & 0.105 & 0.133 & -- & 0.114 & 0.100 \\
\cmidrule(lr){2-14}
& \ngram
& 0 & 0 & 0 & -- & 0 & 0
& 0 & 0 & 0 & -- & 0 & 0 \\
& \xlmroberta
& \textbf{0.947} & \textbf{0.866} & \textbf{0.926} & -- & \textbf{0.836} & \textbf{0.756}
& \textbf{0.753} & \textbf{0.914} & \underline{0.857} & -- & \textbf{0.949} & \underline{0.622} \\
& \detective
& \underline{0.300} & \underline{0.266} & \underline{0.483} & -- & \underline{0.200} & \underline{0.434}
& \underline{0.681} & \underline{0.867} & \textbf{0.933} & -- & \underline{0.847} & \textbf{0.681} \\
\cmidrule(lr){2-14}
& \tracerank
& 0.128 & 0.131 & 0.080 & -- & 0.146 & 0
& 0.157 & 0.170 & 0.174 & -- & 0 & 0 \\
& \traceentjs
& 0 & 0 & 0 & -- & 0 & 0
& 0 & 0 & 0 & -- & 0 & 0 \\
& \traceentnor
& 0 & 0 & 0 & -- & 0 & 0
& 0 & 0 & 0 & -- & 0 & 0 \\

\midrule

\multirow{9}{*}{\rus{}}
& \rank
& 0.05 & 0.068 & 0.085 & 0.064 & -- & 0.069
& 0.070 & 0.068 & 0.061 & 0.075 & -- & 0.096 \\
& \entropy
& 0.051 & 0.057 & 0.148 & 0.064 & -- & \underline{0.190}
& 0.034 & 0.061 & 0.106 & 0.046 & -- & 0.072 \\
& \gltr
& 0.063 & 0 & 0 & 0.076 & -- & 0.094
& 0.07 & 0.095 & \textbf{0.2} & 0.106 & -- & \underline{0.229} \\
\cmidrule(lr){2-14}
& \ngram
& 0 & 0 & 0 & 0 & -- & 0
& 0 & 0 & 0 & 0 & -- & 0 \\
& \xlmroberta
& \textbf{0.482} & \textbf{0.695} & \textbf{0.710} & \textbf{0.749} & -- & \textbf{0.393}
& \textbf{0.298} & \textbf{0.531} & \textbf{0.419} & \textbf{0.380} & -- & \textbf{0.420} \\
& \detective
& 0.063 & 0.283 & 0.114 & \underline{0.250} & -- & 0.146
& \underline{0.200} & \underline{0.200} & \underline{0.200} & \underline{0.184} & -- & 0 \\
\cmidrule(lr){2-14}
& \tracerank
& \underline{0.125} & \underline{0.413} & \underline{0.305} & 0.214 & -- & 0
& 0 & 0.270 & 0 & 0 & -- & 0 \\
& \traceentjs
& 0 & 0 & 0 & 0 & -- & 0
& 0 & 0 & 0 & 0 & -- & 0 \\
& \traceentnor
& 0 & 0.142 & 0 & 0 & -- & 0
& 0 & 0 & 0 & 0 & -- & 0 \\

\midrule

\multirow{9}{*}{\zh{}}
& \rank
& 0.070 & 0.068 & 0.055 & 0.053 & 0.061 & --
& 0.080 & 0 & 0 & 0 & 0 & -- \\
& \entropy
& 0.070 & 0.196 & 0.082 & 0.048 & 0.074 & --
& 0.070 & 0.150 & 0.055 & 0.075 & 0.074 & -- \\
& \gltr
& 0.055 & 0.105 & 0.077 & 0.066 & 0.074 & --
& 0.085 & 0 & 0 & 0 & 0.066 & -- \\
\cmidrule(lr){2-14}
& \ngram
& 0.080 & 0.068 & 0.077 & 0.075 & 0.061 & --
& 0 & 0 & 0 & 0 & 0 & -- \\
& \xlmroberta
& \textbf{0.717} & \textbf{0.871} & \textbf{0.781} & \textbf{0.797} & \textbf{0.950} & --
& \underline{0.315} & \textbf{0.711} & \textbf{0.719} & \textbf{0.715} & \textbf{0.683} & -- \\
& \detective
& \underline{0.673} & \underline{0.667} & \underline{0.520} & \underline{0.667} & \underline{0.629} & --
& \textbf{0.575} & \underline{0.550} & \underline{0.433} & \underline{0.360} & \underline{0.346} & -- \\
\cmidrule(lr){2-14}
& \tracerank
& 0 & 0.080 & 0 & 0 & 0.05 & --
& 0.023 & 0.109 & 0 & 0 & 0 & -- \\
& \traceentjs
& 0 & 0 & 0 & 0 & 0 & --
& 0 & 0 & 0 & 0 & 0 & -- \\
& \traceentnor
& 0 & 0 & 0 & 0 & 0 & --
& 0 & 0 & 0 & 0 & 0 & -- \\
\bottomrule[1.5pt]
\end{tabular}%
}
\caption{
Threshold-based \oodThree results across source and target languages
under low- and high-resource settings. Rows indicate the source language
used for training, while columns indicate the target language used for
evaluation.
}
\label{tab:ood_language_threshold_all}
\end{table*}

\begin{table*}[p]
\centering
\small
\resizebox{\textwidth}{!}{%
\begin{tabular}{llccccccc@{\hspace{0.45cm}}ccccccc}
\toprule[1.5pt]
\multirow{2}{*}{\textbf{Train}} 
& \multirow{2}{*}{\textbf{Method}}
& \multicolumn{6}{c}{\textbf{Low Resource}}
& \multicolumn{6}{c}{\textbf{High Resource}} \\
\cmidrule(lr){3-8}
\cmidrule(lr){9-14}
& 
& \textbf{\en{}} & \textbf{\de{}} & \textbf{\ita{}} & \textbf{\es{}} & \textbf{\rus{}} & \textbf{\zh{}}
& \textbf{\en{}} & \textbf{\de{}} & \textbf{\ita{}} & \textbf{\es{}} & \textbf{\rus{}} & \textbf{\zh{}} \\
\midrule

\multirow{9}{*}{\en{}}
& \rank
& -- & 0.068 & 0.055 & 0.053 & 0.061 & 0.071
& -- & 0.068 & 0.055 & 0.075 & 0.074 & 0.082 \\
& \entropy
& -- & 0.068 & 0.055 & 0.053 & 0.061 & 0.071
& -- & 0.057 & 0.055 & 0.053 & 0.061 & 0.071 \\
& \gltr
& -- & 0.138 & 0.077 & 0.064 & 0.074 & 0.059
& -- & 0.283 & 0.055 & 0.157 & 0.074 & 0.228 \\
\cmidrule(lr){2-14}
& \ngram
& -- & 0.068 & 0.16 & 0.064 & 0.156 & 0.059
& -- & 0.764 & 0.490 & 0.533 & 0.361 & 0.139 \\
& \xlmroberta
& -- & \textbf{0.711} & \textbf{0.685} & \textbf{0.705} & \textbf{0.696} & \textbf{0.380}
& -- & \textbf{0.916} & \textbf{1} & \textbf{1} & \textbf{0.955} & \textbf{0.482} \\
& \detective
& -- & \underline{0.483} & \underline{0.474} & \underline{0.466} & \underline{0.448} & \underline{0.210}
& -- & \underline{0.868} & \underline{0.846} & \underline{0.834} & \underline{0.707} & \underline{0.348} \\
\cmidrule(lr){2-14}
& \tracerank
& -- & 0.168 & 0.404 & 0.298 & 0.300 & 0.08
& -- & 0.130 & 0.298 & 0.172 & 0.342 & 0.137 \\
& \traceentjs
& -- & 0.071 & 0.386 & 0.144 & 0.220 & 0.144
& -- & 0.281 & 0.08 & 0.435 & 0.245 & 0.361 \\
& \traceentnor
& -- & 0.409 & 0.209 & 0.295 & 0.331 & 0.240
& -- & 0.076 & 0.077 & 0.4 & 0.119 & 0.244 \\

\midrule

\multirow{9}{*}{\de{}}
& \rank
& 0.070 & -- & 0.066 & 0.118 & 0.061 & 0.059
& 0.070 & -- & 0.055 & 0.075 & 0.074 & 0.082 \\
& \entropy
& 0.068 & -- & 0.055 & 0.075 & 0.074 & 0.082
& 0.032 & -- & 0.055 & 0.075 & 0.074 & 0.082 \\
& \gltr
& 0.070 & -- & 0.176 & 0.175 & 0.236 & 0.085
& 0.070 & -- & 0.055 & 0.075 & 0.074 & 0.082 \\
\cmidrule(lr){2-14}
& \ngram
& 0.364 & -- & \underline{0.560} & 0.394 & 0.23 & 0.059
& 0.549 & -- & 0.819 & 0.842 & 0.618 & 0.141 \\
& \xlmroberta
& \underline{0.536} & -- & 0.483 & \underline{0.591} & 0.602 & \textbf{0.501}
& \underline{0.826} & -- & \textbf{0.959} & \textbf{1} & \textbf{0.916} & \textbf{0.871} \\
& \detective
& \textbf{0.633} & -- & \textbf{0.697} & \textbf{0.681} & \underline{0.603} & 0.301
& \textbf{0.840} & -- & \underline{0.878} & \underline{0.962} & 0.488 & \underline{0.668} \\
\cmidrule(lr){2-14}
& \tracerank
& 0.060 & -- & 0.060 & 0.384 & 0.258 & 0.109
& 0.155 & -- & 0.225 & 0.337 &\underline{0.704} & 0.226 \\
& \traceentjs
& 0.271 & -- & 0.438 & 0.290 & \textbf{0.774} & 0.304
& 0.264 & -- & 0.375 & 0.266 & 0.544 & 0.355 \\
& \traceentnor
& 0.225 & -- & 0.148 & 0.114 & 0.532 & 0.360
& 0.276 & -- & 0.366 & 0.391 & 0.488 & 0.264 \\

\midrule

\multirow{9}{*}{\ita{}}
& \rank
& 0.050 & 0.068 & -- & 0.064 & 0.144 & 0.069
& 0.050 & 0.071 & -- & 0.179 & 0.064 & 0.094 \\
& \entropy
& 0.070 & 0.083 & -- & 0.045 & 0.094 & 0.466
& 0.070 & 0.095 & -- & 0.078 & 0.074 & 0.285 \\
& \gltr
& 0.050 & 0.068 & -- & 0.064 & 0.074 & 0.059
& 0.248 & 0.172 & -- & 0.313 & 0.088 & 0.260 \\
\cmidrule(lr){2-14}
& \ngram
& 0.100 & 0.068 & -- & 0.075 & 0.061 & 0.059
& 0.962 & 0.667 & -- & \underline{0.963} & \textbf{0.902} & 0.175 \\
& \xlmroberta
& \textbf{0.772} & \textbf{0.696} & -- & \textbf{0.760} & \textbf{0.772} & \textbf{0.744}
& \underline{0.938} & \textbf{0.916} & -- & \textbf{1} & 0.783 & \textbf{0.849} \\
& \detective
& \underline{0.431} & \underline{0.432} & -- & 0.445 & 0.272 & \underline{0.290}
& \textbf{0.959} & \textbf{0.916} & -- & 0.941 & \underline{0.871} & \underline{0.832} \\
\cmidrule(lr){2-14}
& \tracerank
& 0.168 & 0.095 & -- & 0.146 & 0.211 & 0.070
& 0.370 & 0.124 & -- & 0.281 & 0.04 & 0.080 \\
& \traceentjs
& 0.163 & 0.418 & -- & \underline{0.460} & 0.34 & 0.238
& 0.344 & \underline{0.405} & -- & 0.286 & 0.394 & 0.144 \\
& \traceentnor
& 0.173 & 0.243 & -- & 0.269 & \underline{0.306} & 0.225
& 0.153 & 0.327 & -- & 0.426 & 0.340 & 0.059 \\

\midrule

\multirow{9}{*}{\es{}}
& \rank
& 0.070 & 0.068 & 0.055 & -- & 0.074 & 0.082
& 0.070 & 0.068 & 0.055 & -- & 0.074 & 0.082 \\
& \entropy
& 0.071 & 0.068 & 0.055 & -- & 0.061 & 0.096
& 0.071 & 0.068 & 0.076 & -- & 0 & 0.096 \\
& \gltr
& 0.070 & 0.068 & 0.055 & -- & 0.074 & 0.088
& 0.070 & 0.080 & 0.185 & -- & 0.133 & 0.096 \\
\cmidrule(lr){2-14}
& \ngram
& 0.450 & 0.566 & 0.509 & -- & 0.380 & 0.197
& 0.729 & 0.667 & 0.777 & -- & \underline{0.900} & 0.289 \\
& \xlmroberta
& \textbf{0.969} & \textbf{0.837} & \textbf{0.926} & -- & \textbf{0.820} & \textbf{0.706}
& \textbf{0.924} & \textbf{1} & \underline{0.922} & -- & \textbf{1} & \underline{0.597} \\
& \detective
& \underline{0.574} & 0.499 & 0.461 & -- & 0.324 & \underline{0.476}
& \underline{0.700} & \underline{0.916} & \textbf{1} & 0.324 & 0.818 & \textbf{0.700} \\
\cmidrule(lr){2-14}
& \tracerank
& 0.119 & 0.126 & 0.246 & -- & 0.382 & 0.008
& 0.171 & 0.163 & 0.254 & -- & 0.320 & 0.088 \\
& \traceentjs
& 0.275 & 0.512 & 0.317 & -- & \underline{0.398} & 0.269
& 0.238 & 0.513 & 0.310 & -- & 0.381 & 0.226 \\
& \traceentnor
& 0.393 & \underline{0.558} & \underline{0.474} & -- & 0.360 & 0.280
& 0.163 & 0.446 & 0.327 & -- & 0.136 & 0.229 \\

\midrule

\multirow{9}{*}{\rus{}}
& \rank
& 0.05 & 0.068 & 0.077 & 0.064 & -- & 0.059
& 0.070 & 0.068 & 0.055 & 0.075 & -- & 0.082 \\
& \entropy
& 0.051 & 0.057 & 0.148 & 0.064 & -- & 0.190
& 0.0347 & 0.059 & 0.148 & 0.046 & -- & 0.152 \\
& \gltr
& 0.05 & 0.068 & 0.077 & 0.064 & -- & 0.059
& 0.070 & 0.076 & 0.176 & 0.231 & -- & 0.296 \\
\cmidrule(lr){2-14}
& \ngram
& 0.070 & 0.068 & 0.055 & 0.053 & -- & 0.071
& 0.189 & 0.235 & 0.232 & 0.209 & -- & 0.197 \\
& \xlmroberta
& 0.590 & 0.805 & \textbf{0.716} & 0.819 & -- & 0.427
& \underline{0.291} & \underline{0.566} & \textbf{0.678} & \textbf{0.592} & -- & \textbf{0.344} \\
& \detective
& 0.217 & 0.268 & \underline{0.360} & 0.274 & -- & 0.273
& 0.256 & \textbf{0.591} & \underline{0.327} & \underline{0.406} & -- & 0.2 \\
\cmidrule(lr){2-14}
& \tracerank
& 0.166 & 0.413 & 0.305 & 0.284 & -- & 0.239
& \textbf{0.355} & 0.603 & 0.171 & 0.335 & -- & 0.104 \\
& \traceentjs
& 0.246 & 0.525 & 0.247 & 0.245 & -- & 0.196
& 0.070 & 0.420 & 0.248 & 0.053 & -- & \underline{0.229} \\
& \traceentnor
& 0.163 & 0.401 & 0.304 & 0.100 & -- & 0.210
& 0.070 & 0.514 & 0.26 & 0.208 & -- & 0.197 \\

\midrule

\multirow{9}{*}{\zh{}}
& \rank
& 0.070 & 0.068 & 0.055 & 0.053 & 0.061 & --
& 0.070 & 0.068 & 0.055 & 0.075 & 0.074 & -- \\
& \entropy
& 0.070 & 0.236 & 0.077 & 0.064 & 0.074 & --
& 0.070 & 0.138 & 0.055 & 0.075 & 0.074 & -- \\
& \gltr
& 0.060 & 0.226 & 0.077 & 0.064 & 0.074 & --
& 0.070 & 0.206 & 0.156 & 0.146 & 0.135 & -- \\
\cmidrule(lr){2-14}
& \ngram
& 0.080 & 0.068 & 0.077 & 0.075 & 0.061 & --
& 0.070 & 0.189 & 0.186 & 0.172 & 0.133 & -- \\
& \xlmroberta
& \underline{0.811} & \textbf{0.953} & \textbf{0.916} & \textbf{0.860} & \textbf{0.959} & --
& \textbf{0.756} & \textbf{0.809} & \textbf{0.949} & \textbf{0.833} & \textbf{0.800} & -- \\
& \detective
& \textbf{0.856} & \underline{0.741} & \underline{0.843} & \underline{0.853} & \underline{0.875} & --
& \underline{0.351} & \underline{0.561} & \underline{0.544} & \underline{0.621} & \underline{0.712} & -- \\
\cmidrule(lr){2-14}
& \tracerank
& 0.080 & 0.205 & 0.077 & 0.142 & 0.156 & --
& 0.082 & 0.225 & 0.077 & 0.092 & 0.094 & -- \\
& \traceentjs
& 0.093 & 0.286 & 0.149 & 0.173 & 0.292 & --
& 0.235 & 0.223 & 0.149 & 0.225 & 0.209 & -- \\
& \traceentnor
& 0.260 & 0.180 & 0.080 & 0.173 & 0.426 & --
& 0.264 & 0.172 & 0.077 & 0.247 & 0.240 & -- \\
\bottomrule[1.5pt]
\end{tabular}%
}
\caption[\oodThree no-threshold results]{\oodThree results across source and target languages under low- and high-resource settings without thresholding. Rows indicate the source language used for training, while columns indicate the target language used for evaluation.}
\label{tab:ood_language_no_threshold_all}
\end{table*}

\end{document}

%% file: def.tex
\usepackage{stmaryrd}
\usepackage{amsfonts}
\usepackage{amssymb}
\usepackage{amsmath}
\usepackage{mathtools}
\usepackage{bbm}
\usepackage{xspace}
\usepackage{todonotes}
\usepackage{adjustbox}
\usepackage{multicol}
\usepackage{spverbatim}
\usepackage{pifont}

\usepackage{algorithm}
\usepackage{algpseudocode}
\usepackage[most]{tcolorbox}
\tcbuselibrary{listings,breakable,skins}
\usepackage{listings}
\usepackage{makecell} 
\usepackage{fontawesome5}
\usepackage{arydshln} %

\usepackage{arydshln}
\usepackage{tabularx,booktabs,multirow}  %
\usepackage{tablefootnote}
\usepackage{subcaption}
\usepackage{enumerate}
\usepackage{siunitx}
\usepackage{soul} %
\usepackage{pgfplots}
\usepackage[table]{xcolor}
\usepackage{enumitem}

\input{math_commands}

\newcommand{\ngram}{\textsc{n-gram}\xspace}
\newcommand{\bert}{\textsc{bert-aa}\xspace}

\newcommand{\detective}{\textsc{detective}\xspace}
\newcommand{\rank}{\textsc{rank}\xspace}
\newcommand{\entropy}{\textsc{entropy}\xspace}
\newcommand{\gltr}{\textsc{gLTR}\xspace}
\newcommand{\trace}{\textsc{trace}\xspace}

\newcommand{\oodOne}{OOD-Domain\xspace}
\newcommand{\oodTwo}{OOD-Author\xspace}
\newcommand{\oodThree}{OOD-Language\xspace}

\newcommand{\en}{\textsf{EN}\xspace}
\newcommand{\de}{\textsf{DE}\xspace}
\newcommand{\ita}{\textsf{IT}\xspace}
\newcommand{\es}{\textsf{ES}\xspace}
\newcommand{\rus}{\textsf{RU}\xspace}
\newcommand{\zh}{\textsf{ZH}\xspace}

\newcommand{\ourdataset}{\textsc{MultiGhostBench}\xspace}

\newcommand{\deepseek}{\texttt{deepseek-v3.2}\xspace}
\newcommand{\qwen}{\texttt{qwen3-235b}\xspace}
\newcommand{\gptOSS}{\texttt{gpt-oss}\xspace}
\newcommand{\geminiPro}{\texttt{gemini-pro}\xspace}
\newcommand{\geminiFlash}{\texttt{gemini-flash}\xspace}
\newcommand{\gpt}{\texttt{GPT-2}\xspace}
\newcommand{\mgpt}{\texttt{mGPT}\xspace}

\newcommand{\gemma}{\texttt{Gemma}\xspace}

\newcommand{\selfBleu}{Self-\textsc{BLEU}\xspace}

\newcommand{\refapp}[1]{Appendix~\ref{#1}}
\newcommand{\refapptab}[1]{Appendix~Table~\ref{#1}}

\newcommand{\reffig}[1]{Figure~\ref{#1}}

\newcommand{\reftab}[1]{Table~\ref{#1}}

\def\eg{{e.g.,}\xspace}
\def\ie{{i.e.,}\xspace}

\def\etc{{etc.}\xspace}

\newcommand{\tracerank}{\textsc{trace}$_{\texttt{rank-js}}$\xspace}
\newcommand{\traceentjs}{\textsc{trace}$_{\texttt{entr-js}}$\xspace}
\newcommand{\traceentnor}{\textsc{trace}$_{\texttt{entr-norm}}$\xspace}

\newcommand{\xlmroberta}{\textsc{xlm-roberta}\xspace}

\newcommand{\cmark}{\textcolor{green!70!black}{\ding{51}}}   %
\newcommand{\xmark}{\textcolor{red}{\ding{55}}}

%% file: math_commands.tex
\usepackage{amsmath,amsfonts,bm}

\def\eqref#1{equation~\ref{#1}}

\def\1{\bm{1}}

\DeclareMathAlphabet{\mathsfit}{\encodingdefault}{\sfdefault}{m}{sl}
\SetMathAlphabet{\mathsfit}{bold}{\encodingdefault}{\sfdefault}{bx}{n}